\documentclass[final]{cvpr}

\usepackage{times}
\usepackage{epsfig}
\usepackage{graphicx}
\usepackage{amsmath}
\usepackage{amssymb}

\usepackage{atbegshi}

\usepackage{etoolbox}

\newbool{draftmode}

\boolfalse{draftmode}

\relax

\ifbool{draftmode}{

    \PackageWarning{customization/core.sty}{Draft mode is on.}

}{
}

\usepackage{algorithm}

\usepackage[
]{algorithmic}

\usepackage{fixme}

\ifbool{draftmode}{

    \ifdef{\fxsetup}{

        \fxsetup{
            status=draft,
            layout=inline,
            author=,
            theme=color
        }

    }{
}

}{

    \ifdef{\fxsetup}{

        \fxsetup{
            status=final
        }

    }{
}

}

\usepackage{appendix}

\usepackage[
    square,
    comma,
    numbers,
    sort&compress
]{natbib}

\ifdef{\addbibresource}{

\addbibresource{references.bib}
\ExecuteBibliographyOptions{
        sorting=6
    }

}{
}

\ifdef{\setcitestyle}{

    \setcitestyle{}

}{
}

\usepackage{microtype}

\usepackage{siunitx}

\usepackage{textcomp}

\usepackage{textgreek}

\usepackage[user,xr]{zref}

\ifdef{\zxrsetup}{

\ifbool{draftmode}{
}{

        \IfFileExists{main.tex}{

            \IfFileExists{sup.tex}{

\zexternaldocument*[main:]{main}

\zexternaldocument*[sup:]{sup}

            }{
}

        }{
}

    }

}{
}

\usepackage{afterpage}

\usepackage{graphicx}

\usepackage{caption}
\usepackage[
    labelformat=simple
]{subcaption}

\ifdef{\captionsetup}{

\captionsetup[figure]{
font=small,
skip=0.5ex
    }

\captionsetup[table]{
font=small,
skip=0.5ex
    }

}{
}

\ifdef{\thesubfigure}{

\renewcommand{\thesubfigure}{~(\alph{subfigure})}

}{
}

\ifdef{\setlist}{

    \setlist{
topsep=0ex,
partopsep=0ex,
parsep=0ex,
itemsep=0.5ex
    }

}{
}

\usepackage{amsmath}

\DeclareMathOperator*{\argmax}{arg\,max}

\usepackage{amssymb}

\usepackage{amsthm}

\usepackage{bbm}

\usepackage{mathrsfs}

\usepackage{mathtools}

\usepackage{xfrac}

\usepackage{titlesec}

\ifdef{\titlespacing}{

\titlespacing{\section}{0pt}{*.05 }{*.05 }
\titlespacing{\subsection}{0pt}{*.05 }{*.05 }
\titlespacing{\subsubsection}{0pt}{*.05 }{*.05 }
\titlespacing{\paragraph}{0pt}{*.05 }{*.05 }
\titlespacing{\subparagraph}{0pt}{*.05 }{*.05 }

}{
}

\usepackage{setspace}

\usepackage{array}

\usepackage{booktabs}

\usepackage{multirow}

\usepackage{tabu}

\usepackage{tabularx}

\usepackage{tabulary}

\ifdef{\heavyrulewidth}{

\setlength{\heavyrulewidth}{.16em}
\setlength{\lightrulewidth}{.16em}
\setlength{\cmidrulewidth}{.16em}
\setlength{\aboverulesep}{0.1ex}
\setlength{\belowrulesep}{0.1ex}

}{
}

\ifcsdef{tabu}{

\setlength{\tabulinesep}{3pt}

}{
}

\newcommand{\teacherset}{\mathcal{T}}

\newcommand{\classset}{\mathcal{C}}
\newcommand{\classsetall}{\mathcal{C}_0}

\newcommand{\imageset}{\mathcal{I}}

\newcommand{\ratio}{\rho}

\newcommand{\iou}{\Phi}
\newcommand{\policy}{\pi}
\newcommand{\fusion}{f}
\newcommand{\overlap}{\varepsilon}
\newcommand{\unlabeled}{c_0}

\newcommand{\softprediction}{\hat{s}}
\newcommand{\hardprediction}{\hat{y}}

\newcommand{\area}[2]{A_{#1}^{#2}}
\newcommand{\fusedarea}[2]{A_{#1}^{#2}}

\newcommand{\xsrc}{x_{src}}
\newcommand{\ysrc}{y_{src}}
\newcommand{\dataset}{\mathcal{D}}
\newcommand{\xtgt}{x_{tgt}}
\newcommand{\ytgt}{y_{tgt}}

\newcommand{\yfused}{\tilde{y}}
\newcommand{\sfused}{\tilde{s}}
 \usepackage{amsmath, amsthm, amssymb}
\usepackage{cancel}
\usepackage{makecell}
\newtheorem{prop}{Proposition}

\usepackage{bbm}
\usepackage{pdfpages}

\usepackage[pagebackref=true,breaklinks=true,colorlinks,bookmarks=false]{hyperref}

\def\cvprPaperID{20} \def\confYear{CVPR 2021}

\begin{document}

\title{Rethinking Ensemble-Distillation for \\Semantic Segmentation Based Unsupervised Domain Adaptation}

\author{Chen-Hao Chao, Bo-Wun Cheng, and Chun-Yi Lee \\
Elsa Lab, Department of Computer Science, National Tsing Hua University, Hsinchu, Taiwan\\
{\tt\small \{lance$\_$chao, bobcheng15, cylee\}@gapp.nthu.edu.tw}
}

\maketitle

\begin{abstract}
Recent researches on unsupervised domain adaptation (UDA) have demonstrated that end-to-end ensemble learning frameworks serve as a compelling option for UDA tasks. Nevertheless, these end-to-end ensemble learning methods often lack flexibility as any modification to the ensemble requires retraining of their frameworks. To address this problem, we propose a flexible ensemble-distillation framework for performing semantic segmentation based UDA, allowing any arbitrary composition of the members in the ensemble while still maintaining its superior performance. To achieve such flexibility, our framework is designed to be robust against the output inconsistency and the performance variation of the members within the ensemble. To examine the effectiveness and the robustness of our method, we perform an extensive set of experiments on both GTA5$\to$Cityscapes and SYNTHIA$\to$Cityscapes benchmarks to quantitatively inspect the improvements achievable by our method. We further provide detailed analyses to validate that our design choices are practical and beneficial. The experimental evidence validates that the proposed method indeed offer superior performance, robustness and flexibility in semantic segmentation based UDA tasks against contemporary baseline methods.
\end{abstract}
 \vspace{-1.5em}
\begin{figure}[t]
    \includegraphics[width=\linewidth]{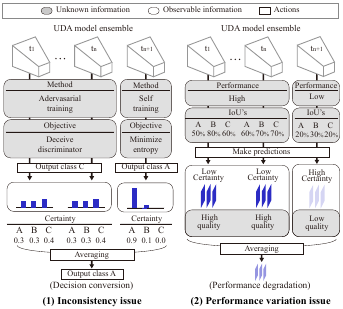}
    \caption{An illustrative example of (1) the inconsistency issue and (2) the performance variation issue mentioned in Section~\ref{sec:introduction}. For (1), the models $t_1,...,t_n$ are trained with ADA methods, which produce relatively low-certainty outputs. In contrast, the model $t_{n+1}$ is trained with self-training method and outputs high-certainty predictions. After averaging, the ensemble predicts class \texttt{A} as its final prediction instead of the majority consensus, i.e., class \texttt{C}. For (2), the models $t_1,...,t_n$ are assumed to be high-performing members, while $t_{n+1}$ is an under-performing one. After averaging, the high-certainty predictions from $t_{n+1}$ could dominate the output of the ensemble, and cause the overall performance to degrade.} 
    \label{fig:issue}
\end{figure} \section{Introduction}
\label{sec:introduction}
In the past few years, semantic segmentation has been attracting the attention of computer vision researchers. Many supervised semantic segmentation methods have been proposed and achieved remarkable performance~\cite{yu2017dilated, lin2017refinenet, yu2015multi, badrinarayanan2017segnet, long2015fully, yuan2019object, wu2019wider, sandler2018mobilenetv2, chen2014semantic, chen2017deeplab, chen2017rethinking, chen2018encoder, zhao2017pyramid}. Typically, those supervised semantic segmentation methods require abundant labeled training data, which are usually expensive to annotate and are commonly unavailable in most real-world scenarios. To resolve this problem, semantic segmentation based unsupervised domain adaptation (UDA) methods~\cite{hoffman2016fcns, hoffman2018cycada, luo2019significance, gong2019dlow, wu2018dcan, tsai2018learning, yang2020adversarial, tsai2019domain, vu2019advent, zhang2018fully, chen2017no, zheng2019unsupervised, choi2019self, zou2018unsupervised, zou2019confidence, zheng2020rectifying, tranheden2020dacs, luo2019taking, lee2018spigan, chen2019learning, vu2019dada, lv2020cross, yang2020fda} have been introduced to bridge different domains. These semantic segmentation based UDA models learn to generalize to a target domain by training with the annotated data from a source domain and the unlabeled data from a target domain. Among these works, the authors in~\cite{luo2019taking, hoffman2016fcns, hoffman2018cycada, luo2019significance, gong2019dlow, wu2018dcan, tsai2018learning, yang2020adversarial, tsai2019domain, vu2019advent, zhang2018fully, chen2017no, zheng2019unsupervised, chen2019crdoco, li2019bidirectional, du2019ssf} resorted to adversarial domain adaptation (ADA) methods, through which the domain discrepancy is minimized by using their adversarial training schemes. Another branch of works has opted for self-training frameworks~\cite{choi2019self, zou2018unsupervised, zou2019confidence, zheng2020rectifying, tranheden2020dacs}, which aim to improve the stability of their models during deployment by minimizing the entropy of the models' predictions in a target domain. These ADA and self-training methods have demonstrated how a single model is able to learn to generalize to an annotation-less target domain. However, they only learn from a single distribution, leaving space for further improvements. Recently, in light of the potential benefits of combining multiple UDA models, a number of works~\cite{nguyen2021unsupervised, kang2019contrastive} have attempted to borrow the concepts from ensemble learning. These works demonstrated how a group of UDA models can be trained simultaneously in an end-to-end fashion to learn different distributions of semantic information, and meanwhile transferring the knowledge to a compact student model. Despite their successes in bridging the domain gaps with multiple learners, these ensemble learning methods often lack flexibility as any modification to the teacher ensemble requires complete retraining of the whole framework.
\vspace{-0.2em}

To address such a problem, the concept of ensemble-distillation~\cite{bucilua2006model, hinton2015distilling, cho2019efficacy, furlanello2018born, balan2015bayesian, nguyen2020joint, orbes2019knowledge, liu2019structured, malinin2019ensemble} can be leveraged since its focus is on designing an effective distillation process instead of a costly end-to-end ensemble learning framework. Typically, these ensemble distillation frameworks view the members in an ensemble as probabilistic models, and transfer the knowledge using expected certainty outputs. Nonetheless, the robustness of these methods is not guaranteed as they do not carefully take into account the followings:
(1) the inconsistency in the scale of the output certainty values among the members in an ensemble (abbreviated as the `\textbf{inconsistency issue}' hereafter), and (2) the performance variations across the members in an ensemble (abbreviated as the `\textbf{performance variation issue}' hereafter). An example of these two issues is illustrated in Fig.~\ref{fig:issue}. For the former, since each teacher model in the ensemble can be trained independently using different methods (e.g., ADA, self-training, data augmentation, or compound usage of them), the scale of the output certainty values may not be consistent across the ensemble. This may result in a situation that few members' decisions with high certainty values dominate the entire ensemble's output. As a result, the outputs from the members in an ensemble should be treated in an equal manner, as the inconsistency in their certainty values may come from their different training objectives instead of the real data distribution in the target domain. For the latter, since the performance (either per-class or average accuracy) of each teacher model in the ensemble may vary substantially, few under-performing members in the ensemble may cause the quality of the combined prediction to degrade significantly. This problem is especially severe under the context of UDA, since the ground truth labels in the target domain are unavailable, and the performance of the ensemble in the target domain is actually unknown.  The above observations suggest that an effective mechanism is necessary to deal with these two issues and prevent them from influencing the quality of the combined predictions.

Being aware of these problems, we introduce a novel ensemble-distillation framework to avoid the aforementioned pitfalls. First, to tackle the certainty inconsistency issue, we introduce an output unification method in the framework to reduce the impact of the inconsistent scales of the certainty outputs. Next, we embrace a new category of fusion function in our framework, named channel-wise fusion, to resolve the performance variation issue. Moreover, we design a method to determine the fusion policy of the proposed channel-wise fusion function to further enhance its effectiveness. To validate our designs, we evaluate the proposed framework with two commonly-adopted metrics, GTA5~\cite{richter2016playing}$\to$Cityscapes~\cite{Cordts2016Cityscapes} and SYNTHIA~\cite{ros2016synthia}$\to$Cityscapes, to demonstrate the effectiveness and robustness of our framework against a number of baselines. The contributions are summarized as follows:
\vspace{-0.5ex}
\begin{itemize}
\setlength\itemsep{-0.7ex}
  \item We introduce a flexible UDA ensemble-distillation framework which is robust against the inconsistency in the scale of the output certainty values and the performance variations among the members in an ensemble.
  \item We propose a new category of fusion function, called channel-wise fusion, along with a fusion policy selection strategy as well as a conflict resolving mechanism to enhance its effectiveness. \item We evaluate our framework under various configurations, and demonstrate that it is able to outperform the baselines in terms of its robustness and effectiveness.
\end{itemize}
\vspace{-0.5ex} \section{Related Works}
\label{sec:related_works}

\paragraph{Unsupervised Domain Adaptation:}~A number of methods have been proposed to bridge the discrepancy between different domains. One branch of these works adopted ADA frameworks to learn representations of their target domains~\cite{luo2019taking, hoffman2016fcns, hoffman2018cycada, luo2019significance, gong2019dlow, wu2018dcan, tsai2018learning, yang2020adversarial, tsai2019domain, vu2019advent, zhang2018fully, chen2017no, zheng2019unsupervised, chen2019crdoco, li2019bidirectional, du2019ssf}. These approaches typically employ a generator and a discriminator trained against each other to minimize the domain gap, and have shown significant improvements over those trained directly in the source domains. Another line of works has turned their attention to self-training and data augmentation measures to tackle UDA problems. For those works utilizing self-training, the concentration was mainly on preventing overfitting by using regularization~\cite{zou2019confidence, zheng2020rectifying} or class-balancing~\cite{zou2018unsupervised} when minimizing the uncertainty in their target domains. The authors in~\cite{tranheden2020dacs} extended the concept of self-training and proposed a data augmentation technique. Their proposed method fine-tunes a model with mixed labels generated by combining ground truth annotations from a source domain and pseudo labels from a target domain. Recent researchers employed ensemble learning frameworks to resolve UDA problems~\cite{nguyen2021unsupervised, kang2019contrastive}. The authors in~\cite{kang2019contrastive} proposed an end-to-end ensemble framework to solve UDA classification problems. The authors in~\cite{nguyen2021unsupervised} extended the idea of ensemble learning and proposed a joint learning ensemble framework to solve person re-identification UDA problems. These works showed how the ensemble learning frameworks can be integrated into UDA.

\paragraph{Pseudo Labeling:}~Pseudo labeling is a self-training method originally proposed to improve the performance of classification networks~\cite{lee2013pseudo}, and is usually accomplished by minimizing the entropy of a model's predictions on unseen data. Pseudo labeling enables a better decision boundary to be achieved as the certainty of a model's prediction increases~\cite{lee2013pseudo}. This concept has been further extended to the field of semantic segmentation, and has gained success by incorporating the information of unlabeled data. Since self-training via pseudo labeling and UDA share many similar characteristics in terms of their problem formulations, it has recently been used to solve UDA problems~\cite{zou2019confidence, zheng2020rectifying, tranheden2020dacs}.

\paragraph{Ensemble-Distillation Method:}~Ensemble-distillation is an extension of knowledge distillation. The authors in~\cite{hinton2015distilling,balan2015bayesian} studied how the knowledge of a teacher model ensemble can be transferred to a student by training it with the soft predictions of the ensemble. They adopted averaging operation for combining the predictions and used KL-divergence as the loss function to transfer the knowledge. The authors in~\cite{malinin2019ensemble} aimed at resolving the diversity collapse issue in the ensemble-distillation problem. They argued that the averaging operation harms the diversity of the models in an ensemble and proposed to use a prior network~\cite{malinin2018predictive} to estimate the distributions of their output uncertainties. These works have demonstrated their effectiveness under supervised training settings. However, the existing ensemble-distillation methods are not designed to handle unsupervised tasks, and are susceptible to the issues introduced in Section~\ref{sec:introduction}. 
\section{Preliminary}
\label{sec:preliminary}
\paragraph{Problem Definition:}~For semantic segmentation based UDA problems, a model has access to the image-label pairs, $\xsrc,\ysrc$, from a source domain dataset $\dataset_{src}$, but only the images $\xtgt$ from a target domain dataset $\dataset_{tgt}$. The training objective is to train the model such that its predictions can best estimate the ground truth labels $\ytgt$ in the target domain. In other words, the mean intersection-over-union (mIoU) between the predictions of the model and $\ytgt$ should be maximized. In the problem formulation concerned by this paper, a pretrained model ensemble $\teacherset$ is given, where each member in $\teacherset$ is separately trained using any arbitrary semantic segmentation based UDA method. The goal is to develop an ensemble-distillation strategy that can effectively integrate the knowledge from $\teacherset$ and distill it into a single student model, in a way that the mIoU of the student's predictions for the instances in $\dataset_{tgt}$ is maximized.

\begin{figure}[t]
\includegraphics[width=\linewidth]{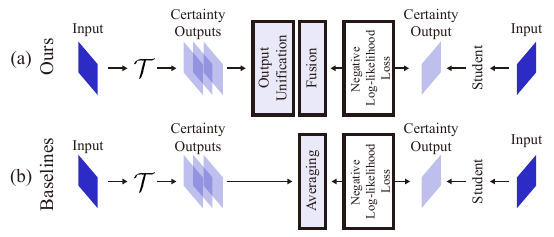}
    \caption{A comparison between (a) the proposed ensemble-distillation framework and (b) the baseline framework.}
    \label{fig:distillation}
\end{figure}

\paragraph{Previous Ensemble-Distillation Method:}~In this section, we explain how the concepts of the previous ensemble-distillation works~\cite{bucilua2006model, hinton2015distilling, balan2015bayesian, cho2019efficacy, furlanello2018born, nguyen2020joint, orbes2019knowledge, liu2019structured} can be borrowed to perform semantic segmentation based UDA ensemble-distillation tasks. Typically, these works view $\teacherset$ as a set of probabilistic models, and complete the ensemble-distillation process through minimizing the negative log-likelihood loss $\mathcal{L}_{KL}$ between the expected outputs from the ensemble and the student model, as depicted in Fig.~\ref{fig:distillation}~(b). The distillation process of these methods under the settings of semantic segmentation based UDA can be formulated as:
\begin{equation}
\label{eq:loss_KL}
    \mathcal{L}_{KL} = -\sum_{p\in \imageset}{\sum_{c\in \classset}{\sfused^{(p,c)}\,\text{log}(r^{(p,c)})}},
\end{equation} where $\imageset$ is a set of pixels in an image, $\classset$ is a given set of semantic classes. $r^{(p,c)}\in \mathbb{R}^{|\imageset|\times|\classset|}$ is the student's certainty output, and $\sfused^{(p,c)}\in \mathbb{R}^{|\imageset|\times|\classset|}$ represents the expected probabilistic prediction for class $c \in \classset$ at pixel $p \in \imageset$. A common way to capture $\sfused$ is through averaging, expressed as follows:
\begin{equation}
\label{eq:certainty}
    \sfused^{(p,c)} = \dfrac{1}{|\teacherset|}\sum_{t\in \teacherset} 
    \softprediction^{(p,c,t)},
\end{equation}
 where $\softprediction^{(p,c,t)}\in \mathbb{R}^{|\imageset|\times|\classset|\times |\teacherset|}$ is the probabilistic prediction from $t \in \teacherset$ for class $c \in \classset$ at pixel $p \in \imageset$ on target instances. As a result, the knowledge of the teacher ensemble can be transferred through minimizing $\mathcal{L}_{KL}$ in a target domain.

\paragraph{Pitfalls:}~As discussed in Section~\ref{sec:introduction}, directly adopting previous methods to solve semantic segmentation based UDA problems is problematic because of the inconsistency issue and the performance variation issue. 
For the former, the scale of $\softprediction$ may vary across the models in $\teacherset$ under our problem formulation. This suggests that a direct operation, such as the averaging operation in Eq.~(\ref{eq:certainty}), is inappropriate as a few $\softprediction$ with high certainty values may dominate the ensemble's output decision. For the performance variation issue, the per-class or the average performance of each member in $\teacherset$ can vary substantially under our problem formulation. Since the fusion function formulated in Eq.~(\ref{eq:certainty}) fuses the predictions of \textit{all} teacher models, the under-performing members in $\teacherset$ can influence the quality of the fused results. Therefore, the adoption of such a fusion function is inappropriate as it may be sensitive to the performance variations within $\teacherset$.

 \section{Methodology}
\label{sec:methodology}
To address the aforementioned problems, we introduce a new ensemble-distillation framework, and illustrate it in Fig.~\ref{fig:distillation}~(a). The main difference between the proposed method and the previous ones lies in two aspects: \textit{Output Unification} and \textit{Fusion Function}. In Sections~\ref{sec:method:output_unification} and~\ref{sec:method:fusion_function}, we walk through the designs of the output unification operation as well as the fusion function, and discuss how they may contribute to resolving the aforementioned issues. Finally, in Section~\ref{sec:method:ensemble_distillation}, we formulate and summarize the proposed ensemble-distillation framework.

\subsection{Output Unification}
\label{sec:method:output_unification}
To resolve the inconsistency issue, we argue that the soft predictions $\softprediction$ in Eq.~(\ref{eq:certainty}) should be unified first in the target domain, as illustrated in Fig.~\ref{fig:distillation}.  This additional unification operation ensures that the raw output certainty values from the models in $\teacherset$ do not directly influence the subsequent fusion results. To accomplish this, we unify the soft predictions by converting them to pseudo labels, so as to make them all bear the same scale, i.e., representing the final decisions of the models in $\teacherset$. The unification operation is formulated as:
\begin{equation}
\label{eq:unification}
\hardprediction^{(p,c,t)} =
\begin{cases}
    1,   & \text{if }  c=\underset{c\in \classset}{\argmax}\,\{\softprediction^{(p,c,t)}\}\\
    0,   & \text{otherwise} 
\end{cases},
\end{equation} where $\hardprediction^{(p,c,t)}$ is the unified output prediction from $t \in \teacherset$ for class $c \in \classset$ at pixel $p \in \imageset$ in the target domain. This operation ensures that the subsequent fusion function can operate on items with a consistent scale, and thus eliminates the impact of the inconsistency in the original certainty outputs.

\begin{figure}[t]
    \includegraphics[width=\linewidth]{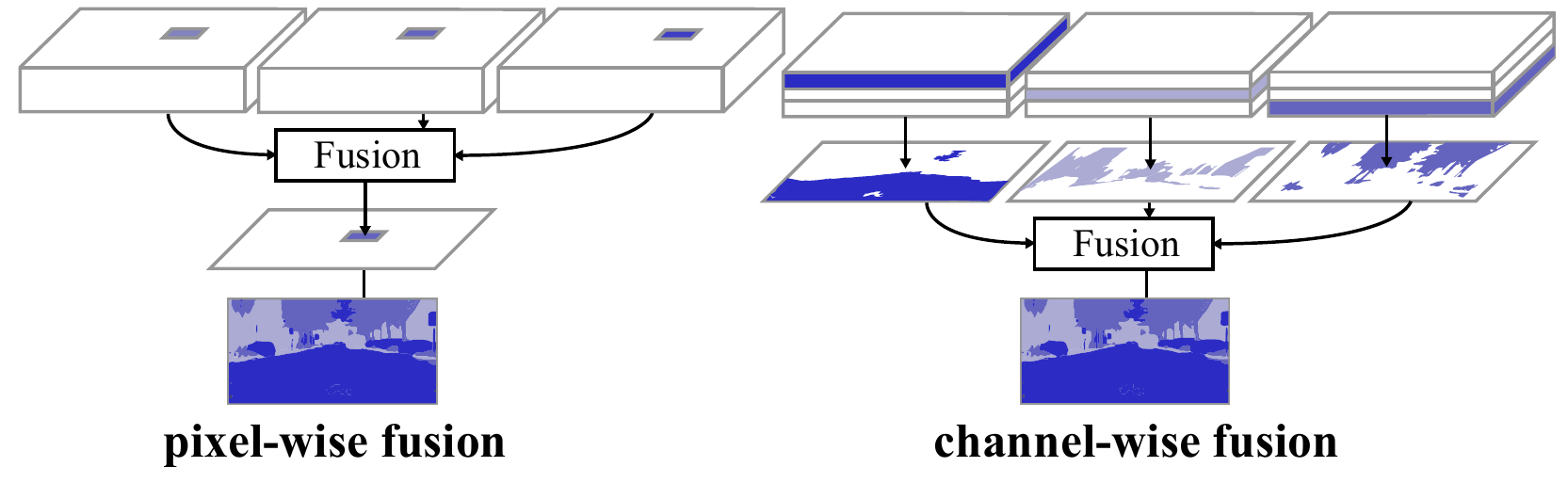}
    \caption{A illustration of the pixel-wise and channel-wise fusions.}
    \label{fig:channel_wise}
\end{figure}
 \subsection{Fusion Function}
\label{sec:method:fusion_function}
We next move on to focus on investigating a fusion function that can take advantage of the unified predictions to achieve  robustness against the performance variations issue. We compare two categories of fusion functions: \textit{pixel-wise fusion} and \textit{channel-wise fusion}. The former is a direct conversion from Eq.~(\ref{eq:certainty}) and is used as our baseline method. The latter is the proposed method and is adopted to address the performance variation issue. Both pixel-wise fusion and channel-wise fusion are mapping functions $\fusion:\imageset\to \classsetall$ that assign a class label $c\in \classsetall$ for the pixel $p\in \imageset$ in the fusion output based on $\hardprediction^{(p,c,t)}$, where $\classsetall := \classset \cup \{\unlabeled\}$ is a set that includes all $c \in \classset$ as well as the unlabeled symbol $\unlabeled$.

\subsubsection{Pixel-Wise Fusion}
Pixel-wise fusion ($\fusion^{Pixel}$) adopts a statistical view on $\teacherset$, and is designed to capture the average behavior of the ensemble. As depicted in Fig.~\ref{fig:channel_wise}, pixel-wise fusion treats each pixel in a semantic segmentation map as the basic unit of the fusion operation. Specifically, the fused result of each pixel is determined by taking majority voting among the predictions from $\teacherset$, and is implemented as the following:

\begin{equation}
\label{eq:majority}
    \fusion^{Pixel}(p) = \underset{c\in \classset}{\argmax}\sum_{t\in \teacherset}{\hardprediction^{(p,c,t)}},
\end{equation} where $\hardprediction$ is the unified output generated according to Eq.~(\ref{eq:unification}).

\begin{figure}[t]
    \includegraphics[width=\linewidth]{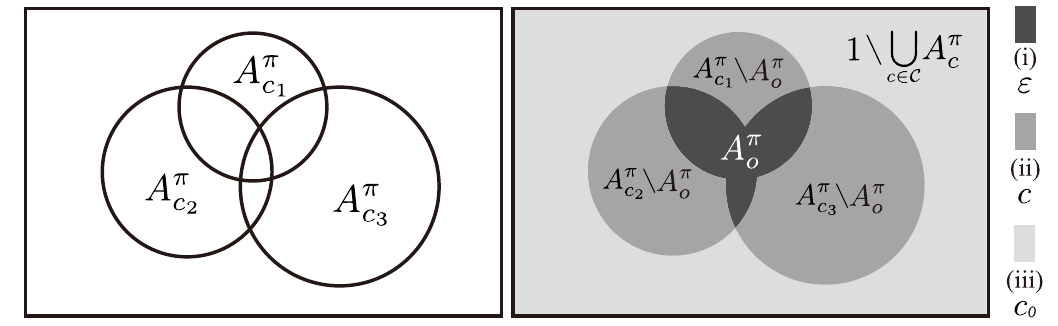}
    \caption{An illustrative example of the three scenarios in Eq.~(\ref{eq:channel_wise}).}
    \label{fig:set}
\end{figure}
 \begin{figure}[t]
    \includegraphics[width=\linewidth]{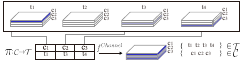}
    \caption{An illustrative example of $\policy$ used in $\fusion^{Channel}$.}
    \label{fig:policy}
\end{figure}

 \subsubsection{Channel-Wise Fusion}
Based on a different perspective, channel-wise fusion ($\fusion^{Channel}$) treats each class channel as the basis for fusion, as depicted in Fig.~\ref{fig:channel_wise}. Instead of fusing the outputs of all $t\in \teacherset$, channel-wise fusion relies on a fusion policy $\policy: \classset \to \teacherset$, which is a mapping function for recombining the unified outputs from different teacher models. More specifically, for each class $c\in \classset$, the fusion policy $\policy$ selects that class channel from the unified output $\hardprediction^{(p,c,t)}$ of a teacher model $t\in \teacherset$, as illustrated in the example shown in Fig.~\ref{fig:policy}.
Given such a $\policy$, the channel-wise fusion function is formulated as:
\begin{equation}
\label{eq:channel_wise}
\fusion^{Channel}(p) =
\begin{cases}
    \text{ (i) }\, \overlap,   & \text{if } p \in \fusedarea{o}{\policy}\\
    \text{(ii) }\,c,          & \text{if } p \in \fusedarea{c}{\policy} \setminus \fusedarea{o}{\policy},\\
    \text{(iii)}\,\unlabeled,   & \text{otherwise} 
\end{cases}
\end{equation}
 where (i) is the condition that $p$ is labeled by multiple teachers, (ii) is the condition that $p$ labeled as a certain class by a single teacher, and (iii) is the condition that $p$ is unlabeled, as illustrated in Fig.~\ref{fig:set}. In Eq.~(\ref{eq:channel_wise}), $\overlap$ denotes a class label to be assigned in scenario (i), $c$ is a class in $\classset$ in scenario (ii), and $\unlabeled$ denotes the unlabeled symbol in scenario (iii). $\fusedarea{c}{\policy}:=\{p\,|\,p\in \imageset,\policy(c)=t,\hat{y}^{(p,c,t)}=1\}$ is a set of pixels comprising  $\hardprediction^{(p,c,t)}$ for a given class $c$ generated by a teacher $t$ selected according to the given fusion policy $\policy$. Since different $\fusedarea{c}{\policy}$ may be produced by different $t\in \teacherset$ for different $c\in \classset$, the pixels they cover are not necessarily mutually exclusive. Therefore, $\fusedarea{o}{\policy}$ is defined to represent the set of pixels labeled by multiple teachers, expressed as follows:
\begin{spacing}{0.7}
\begin{equation}
\label{eq:overlap_area}
    \fusedarea{o}{\policy} = \bigcup_{ \substack{c_{1} \neq c_{2}, \\ c_{1}, c_{2} \in \classset} }{(\fusedarea{c_1}{\policy} \cap \fusedarea{c_2}{\policy})}.
\end{equation}
\end{spacing} The mechanism that assigns the value of $\overlap$ for all $p \in \fusedarea{o}{\policy}$ is referred as the conflict-resolving mechanism. In this work, we employ a spatially-aware conflict-resolving mechanism that assigns a class label for each pixel in $\fusedarea{o}{\policy}$ using majority voting on a kernel. The size of the kernel is denoted as $\kappa$, and the set of pixels covered by the kernel centered at $p$ is referred to as $B^{\kappa}_p$. The mechanism is formulated as follows:
\vspace{-0.15em}
\begin{equation}
\label{eq:fused_area}
    \overlap = \underset{c\in \classset^{\policy}_p}{\argmax}{|B^{\kappa}_p\cap \fusedarea{c}{\policy}|}.
\end{equation}
 where $\classset^{\policy}_p:=\{c\,|\,c \in \classset; p\in \area{c}{\policy}\}$ represents a set of class(es) assigned to a pixel $p$ under $\area{c}{\policy}$.

\subsubsection{Theoretical Properties of Channel-Wise Fusion}
Let $\iou^{(c,t)}$ and $\Tilde{\iou}^{(c,\policy(c))}$ be the per-class IoU's w.r.t. $\ytgt$ for the pseudo labels generated by $t\in \teacherset$ and the fused pseudo labels generated by $f^{Channel}$, respectively. Channel-wise fusion conforms to the following properties:

\begin{prop}
Consider an arbitrary fusion policy $\policy$. Given a constant $\alpha \in (0,1)$ and classes $c_1, ..., c_n\in \classset$. If $\iou^{(c_i,t)} \geq \alpha, \forall i\in \{1, ..., n\},\,\forall t\in \teacherset$ and $|\fusedarea{o}{\policy}|=0$, we have:
\vspace{-0.2em}
\begin{equation}
   \text{mIoU}=\frac{1}{|\classset|}\sum_{c\in \classset} \Tilde{\iou}^{(c,\policy(c))} \geq \frac{n\alpha}{|\classset|}.
\end{equation}
\end{prop}

\vspace{-0.9em}

\begin{prop}
Consider an optimal fusion policy $\policy^{*}(c)=\argmax_{t\in \teacherset}{\{\iou^{(c,t)}\}}$. Assume $|\fusedarea{o}{\policy^{*}}|=0$, we have:
\vspace{-0.2em}
\begin{equation}
   \text{mIoU}=\frac{1}{|\classset|}\sum_{c\in \classset} \Tilde{\iou}^{(c,\policy^{*}(c))} \geq \frac{1}{|\classset|}\sum_{c\in \classset} \iou^{(c,t)}, \forall t\in \teacherset.
\end{equation}
\end{prop}

For the detailed elaborations and proofs with regard to these properties, please refer to the supplementary materials. \textbf{Proposition 1} states the condition when the mIoU lower bound can be ensured. On the other hand, \textbf{Proposition 2} describes how the effectiveness of channel-wise fusion can be maximized. In the next section, we discuss how an effective $\policy$ can be determined.

\begin{figure}[t]
\includegraphics[width=\linewidth]{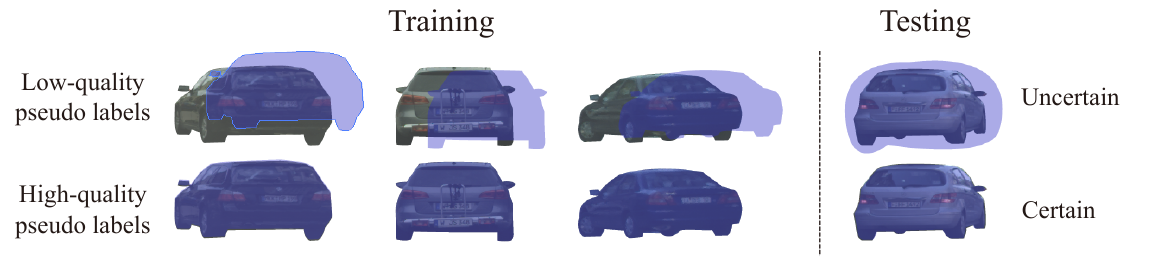}
    \caption{An illustration of how the quality of a teacher model's pseudo labels can affect the output certainty values of the student model. If the student model is trained with high-quality pseudo labels (i.e., pseudo labels with high IoU's w.r.t. $\ytgt$), it can learn a mapping from input features to the segmentation mask effectively, and generates high-certainty predictions. In contrast, if the student model is trained with low-quality pseudo labels (i.e., pseudo labels with low IoU's w.r.t $\ytgt$) that are mismatched with the input features, the student's output certainty values are likely to degrade.}
    \label{fig:compare}
\end{figure}
 \begin{figure}[t]
    \includegraphics[width=\linewidth]{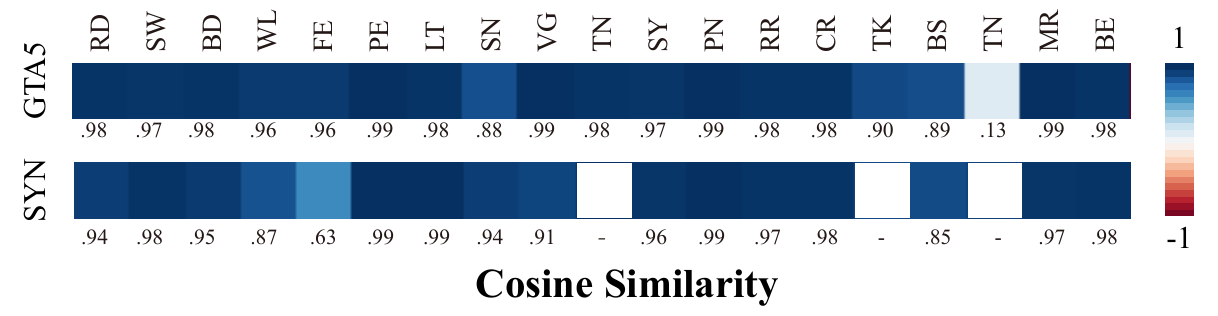}
    \caption{The cosine similarity between the per-class IoU of each $t \in \teacherset$ and the per-class certainty values of the student evaluated on the training set of Cityscapes, i.e., the cosine similarity between $\iou^{(c,t)}$ in Proposition 2 and $\ratio^{(c,t)}$ in Eq.~(\ref{eq:policy}). The experimental results reveal that the two variables are positively correlated to each other, as the cosine similarity values for all $c \in \classset$ are greater than 0 and close to 1. For the detailed settings, please refer to Section~\ref{sec:setup}.}
    \label{fig:similarity}
\end{figure} \subsubsection{Certainty-Aware Policy Selection Strategy}
\label{subsubsec::certainty-aware-policy}
Since the fusion policy $\policy$ determines which teacher model is allowed to involve in $\fusion^{Channel}$, choosing an appropriate $\policy$ is therefore crucial to the robustness of our framework. In order to achieve this objective, a suitable measure is necessary for evaluating the quality of each teacher's predictions in the target domain without using any target domain ground truth. The experimental clue illustrated and explained in Fig.~\ref{fig:compare} and \ref{fig:similarity} offers an empirical manner for the above purpose. From Fig.~\ref{fig:compare}, it is observed that the quality of the unified outputs $\hardprediction^{(p,c,t)}$, i.e., the pseudo labels, from $t\in\teacherset$ are positively correlated to the output certainty values of distilled student model. This correlation suggests that any low-quality $\hardprediction^{(p,c,t)}$, which might be generated by some under-performing teacher models, can confuse the student model and causes its output certainty values to degrade. This correlation between the quality of the pseudo labels from $t\in\teacherset$ and the student's output certainty values thus shed light on the development of the measure for approximating a teacher model's performance without using target domain ground truth. In practice, an offline fusion policy selection strategy is adopted. Our framework first performs knowledge distillation on all $t \in \teacherset$ and transfers their knowledge to $|\teacherset|$ identical student models using the unified outputs. Then, their output certainty values are measured to obtain the approximated performance of their corresponding teacher models. Finally, for each $c \in \classset$, a $t \in \teacherset$ that maximizes the student model's output certainty values is selected. The fusion policy $\policy$ is written as follows:
\begin{equation}
\label{eq:policy}
    \policy(c)=\underset{t\in \teacherset}{\argmax}{\{\ratio^{(c,t)}\}},
\end{equation}
 where $\ratio^{(c,t)} \in \mathbb{R}^{|\classset|\times|\teacherset|}$ refers to the average certainty outputs of class $c\in \classset$ from the student model trained with the unified outputs generated by teacher $t\in \teacherset$.

\subsection{The Proposed Framework}
\label{sec:method:ensemble_distillation}
Based on the formulations of output unification and fusion function, the loss function $\mathcal{L}_{CE}$ for performing the ensemble-distillation in our framework is defined as follows:
\begin{equation}
\label{eq:loss_CE}
    \mathcal{L}_{CE} = -\sum_{p\in \imageset}{\sum_{c\in \classset}{\yfused^{(p,c)}\,\text{log}(r^{(p,c)})}},
\end{equation} where $\yfused^{(p,c)}\in \{0,1\}^{|\imageset|\times|\classset|}$ is the fused results, defined as:
\begin{equation}
\label{eq:pseudo_label}
\yfused^{(p,c)} =
\begin{cases}
    1,   & \text{if }  c=\fusion^{Channel}(p)\\
    0,   & \text{otherwise} 
\end{cases}.
\end{equation} The pseudo code of the ensemble-distillation method in our framework is summarized in the supplementary materials. 

\section{Experimental Setup}
\label{sec:setup}
\paragraph{Baselines and Evaluation Methods:}~In this work, we evaluate and compare the experimental results in terms of the effectiveness and the robustness. To examine the effectiveness of the proposed framework, we compare our method against two ensemble-distillation schemes and a number of UDA baselines. The ensemble-distillation schemes include EnD~\cite{hinton2015distilling} and its recent revision EnD$^2$~\cite{malinin2019ensemble}. The semantic segmentation based UDA baselines cover APODA~\cite{yang2020adversarial}, PatchAlign~\cite{tsai2019domain}, AdvEnt~\cite{vu2019advent}, FDA-MBT~\cite{yang2020fda}, PIT~\cite{lv2020cross}, CBST~\cite{zou2018unsupervised}, MRKLD~\cite{zou2019confidence}, R-MRNet~\cite{zheng2020rectifying}, and DACS~\cite{tranheden2020dacs}. To examine the robustness of our framework, we select the members for $\teacherset$ based on two criteria: (1) each member in $\teacherset$ is trained with different UDA methods; and (2) there exists large per-class and average performance variations among the members in $\teacherset$. According to these criteria, we select DACS~\cite{tranheden2020dacs} (data augmentation), R-MRNet~\cite{zheng2020rectifying} (adversarial training), MRKLD~\cite{zou2019confidence} (self-training), and CBST~\cite{zou2018unsupervised} (self-training) to form $\teacherset$ in our experiments. We evaluate the proposed framework and the baselines on two commonly adopted benchmarks: GTA5~\cite{richter2016playing}$\to$Cityscapes~\cite{Cordts2016Cityscapes} and SYNTHIA~\cite{ros2016synthia}$\to$Cityscapes. For the former, the models have access to 24,966 image-label pairs from the training set of GTA5, and 2,975 images from the training set of Cityscapes. We evaluate the student model's per-class IoU's of the 19 semantic classes as well as the its mIoU's on the validation set of Cityscapes. For the latter, the models have access to 9,400 image-label pairs from the training set of SYNTHIA, and 2,975 images from the training set of Cityscapes. In a similar fashion, we evaluate the student model's per-class IoU's of 13 and 16 semantic classes as well as its mIoU's on the validation set of Cityscapes.

\paragraph{Implementation Details:}~For the student model, we adopt Deeplabv3+~\cite{chen2018encoder} architecture with DRN-D-54~\cite{yu2017dilated} as our backbone, which is trained using SGD with a learning rate initialized to $2.5\times10^{-4}$ and decreased with a factor of $0.9$. The weight decay is set to $5\times10^{-3}$, the momentum is set to 0.9, and the batch size is set to 10 for 100K iterations with early stopping. The value of $\kappa$ in $\fusion^{Channel}$ is set to $13$. During the process of certainty-aware policy selection strategy, 500 images from the training set of Cityscapes are used for measuring $\ratio^{(c,t)}$, while the other 2475 images are used for training the student. The student model is pre-trained in the source domain, and fine-tuned with $(\xsrc,\ysrc)$ and $(\xtgt,\tilde{y})$ during the distillation process.
 \begin{table*}[t]
\centering

\resizebox{\textwidth}{!}{\renewcommand{\arraystretch}{1.11}
\newcommand{\mytoprule}{\toprule[1.5pt]}
\footnotesize
\begin{tabular}{clllllllllllllllllllll}
\mytoprule
\multicolumn{21}{c}{GTA5 $\to$ Cityscapes} &  \\ \mytoprule
\multicolumn{1}{c|}{Method} & Road & SideW & Build & Wall & Fence & Pole & Light & Sign & Veg & Terrain & Sky & Person & Rider & Car & Truck & Bus & Train & Motor & \multicolumn{1}{l|}{Bike} & mIoU &  \\ \hline
\multicolumn{1}{c|}{APODA~\cite{yang2020adversarial}} & 85.6 & 32.8 & 79.0 & 29.5 & 25.5 & 26.8 & 34.6 & 19.9 & 83.7 & 40.6 & 77.9 & 59.2 & 28.3 & 84.6 & 34.6 & 49.2 & 8.0 & 32.6 & \multicolumn{1}{l|}{39.6} & 45.9 &  \\
\multicolumn{1}{c|}{PatchAlign~\cite{tsai2019domain}} & 92.3 & 51.9 & 82.1 & 29.2 & 25.1 & 24.5 & 33.8 & 33.0 & 82.4 & 32.8 & 82.2 & 58.6 & 27.2 & 84.3 & 33.4 & 46.3 & 2.2 & 29.5 & \multicolumn{1}{l|}{32.3} & 46.5 &  \\
\multicolumn{1}{c|}{AdvEnt~\cite{vu2019advent}} & 89.4 & 33.1 & 81.0 & 26.6 & 26.8 & 27.2 & 33.5 & 24.7 & 83.9 & 36.7 & 78.8 & 58.7 & 30.5 & 84.8 & 38.5 & 44.5 & 1.7 & 31.6 & \multicolumn{1}{l|}{32.4} & 45.5 &  \\
\multicolumn{1}{c|}{FDA-MBT~\cite{yang2020fda}} & 92.5 & 53.3 & 82.4 & 26.5 & 27.6 & 36.4 & 40.6 & 38.9 & 82.3 & 39.8 & 78.0 & 62.6 & 34.4 & 84.9 & 34.1 & 53.1 & 16.9 & 27.7 & \multicolumn{1}{l|}{46.4} & 50.5 &  \\
\multicolumn{1}{c|}{PIT~\cite{lv2020cross}} & 87.5 & 43.4 & 78.8 & 31.2 & 30.2 & 36.3 & 39.9 & 42.0 & 79.2 & 37.1 & 79.3 & 65.4 & 37.5 & 83.2 & 46.0 & 45.6 & 25.7 & 23.5 & \multicolumn{1}{l|}{49.9} & 50.6 &  \\ \multicolumn{1}{c|}{\textbf{\textcolor{blue}{CBST}}~\cite{zou2018unsupervised}} & 91.8 & 53.5 & 80.5 & 32.7 & 21.0 & 34.0 & 28.9 & 20.4 & 83.9 & 34.2 & 80.9 & 53.1 & 24.0 & 82.7 & 30.3 & 35.9 & 16.0 & 25.9 & \multicolumn{1}{l|}{42.8} & 45.9 &  \\
\multicolumn{1}{c|}{\textbf{\textcolor{blue}{MRKLD}}~\cite{zou2019confidence}} & 91.0 & 55.4 & 80.0 & 33.7 & 21.4 & 37.3 & 32.9 & 24.5 & 85.0 & 34.1 & 80.8 & 57.7 & 24.6 & 84.1 & 27.8 & 30.1 & \textbf{26.9} & 26.0 & \multicolumn{1}{l|}{42.3} & 47.1 &  \\
\multicolumn{1}{c|}{\textbf{\textcolor{blue}{R-MRNet}}~\cite{zheng2020rectifying}} & 90.4 & 31.2 & 85.1 & 36.9 & 25.6 & 37.5 & 48.8 & 48.5 & 85.3 & 34.8 & 81.1 & 64.4 & 36.8 & 86.3 & 34.9 & 52.2 & 1.7 & 29.0 & \multicolumn{1}{l|}{44.6} & 50.3 &  \\ 
\multicolumn{1}{c|}{\textbf{\textcolor{blue}{DACS}}~\cite{tranheden2020dacs}} & 89.90 & 39.66 & 87.87 & 30.71 & 39.52 & 38.52 & 46.43 & 52.79 & 87.98 & 43.96 & \textbf{88.76} & 67.20 & 35.78 & 84.45 & 45.73 & 50.19 & 0.00 & 27.25 & \multicolumn{1}{l|}{33.96} & 52.14 &  \\ 
\hline \hline
\multicolumn{1}{c|}{Source Only} & 57.40 & 21.43 & 56.80 & 8.93 & 22.14 & 32.38 & 34.62 & 24.90 & 78.98 & 15.92 & 63.71 & 55.55 & 13.83 & 58.11 & 21.99 & 29.78 & 2.36 & 28.41 & \multicolumn{1}{l|}{33.98} & 34.80 &  \\
\multicolumn{1}{c|}{EnD~\cite{hinton2015distilling}} & 92.17 & 53.12 & 84.85 & 24.77 & 29.76 & 40.38 & 40.98 & 49.35 & 86.21 & 42.85 & 79.74 & 62.79 & 35.98 & 85.72 & 42.10 & 44.45 & 0.26 & 28.27 & \multicolumn{1}{l|}{51.80} & 51.34 &  \\ 
\multicolumn{1}{c|}{EnD$^{2}$~\cite{malinin2019ensemble}} & 92.39 & 53.84 & 85.34 & 24.51 & 30.53 & 40.28 & 42.40 & 50.28 & 86.19 & 43.39 & 80.55 & 63.26 & 36.75 & 86.15 & 43.95 & 43.91 & 0.20 & 30.17 & \multicolumn{1}{l|}{53.22} & 51.96 &  \\
\multicolumn{1}{c|}{Ours (Pixel)} & 92.29 & 57.34 & 84.09 & 36.75 & 29.17 & 41.37 & 48.96 & 42.26 & 86.91 & 39.95 & 82.81 & 66.29 & 37.42 & 86.94 & 35.21 & 48.82 & 1.48 & 40.78 & \multicolumn{1}{l|}{53.02} & 53.26 &  \\
\multicolumn{1}{c|}{Ours (Channel)} & \textbf{94.43} & \textbf{60.90} & \textbf{88.07} & \textbf{39.46} & \textbf{41.80} & \textbf{43.24} & \textbf{49.08} & \textbf{56.00} & \textbf{88.01} & \textbf{45.83} & 87.79 & \textbf{67.58} & \textbf{38.05} & \textbf{90.08} & \textbf{57.64} & \textbf{51.90} & 0.00 & \textbf{46.57} & \multicolumn{1}{l|}{\textbf{55.28}} & \textbf{57.98} &  \\ \mytoprule
\multicolumn{21}{c}{SYNTHIA $\to$ Cityscapes} &  \\ \mytoprule
\multicolumn{1}{c|}{Method} & Road & SideW & Build & Wall* & Fence* & Pole* & Light & Sign & Veg & Terrain & Sky & Person & Rider & Car & Truck & Bus & Train & Motor & \multicolumn{1}{l|}{Bike} & mIoU & mIoU* \\ \hline
\multicolumn{1}{c|}{APODA~\cite{yang2020adversarial}} & 86.4 & 41.3 & 79.3 & \multicolumn{1}{c}{-} & \multicolumn{1}{c}{-} & \multicolumn{1}{c}{-} & 22.6 & 17.3 & 80.3 & \multicolumn{1}{c}{-} & 81.6 & 56.9 & 21.0 & 84.1 & \multicolumn{1}{c}{-} & 49.1 & \multicolumn{1}{c}{-} & 24.6 & \multicolumn{1}{l|}{45.7} & \multicolumn{1}{c}{-} & 53.1 \\
\multicolumn{1}{c|}{PatchAlign~\cite{tsai2019domain}} & 82.4 & 38.0 & 78.6 & 8.7 & 0.6 & 26.0 & 3.9 & 11.1 & 75.5 & \multicolumn{1}{c}{-} & 84.6 & 53.5 & 21.6 & 71.4 & \multicolumn{1}{c}{-} & 32.6 & \multicolumn{1}{c}{-} & 19.3 & \multicolumn{1}{l|}{31.7} & 40.0 & 46.5 \\
\multicolumn{1}{c|}{AdvEnt~\cite{vu2019advent}} & 85.6 & 42.2 & 79.7 & 8.7 & 0.4 & 25.9 & 5.4 & 8.1 & 80.4 & \multicolumn{1}{c}{-} & 84.1 & 57.9 & 23.8 & 73.3 & \multicolumn{1}{c}{-} & 36.4 & \multicolumn{1}{c}{-} & 14.2 & \multicolumn{1}{l|}{33.0} & 41.2 & 48.0 \\
\multicolumn{1}{c|}{FDA-MBT~\cite{yang2020fda}} & 79.3 & 35.0 & 73.2 & \multicolumn{1}{c}{-} & \multicolumn{1}{c}{-} & \multicolumn{1}{c}{-} & 19.9 & 24.0 & 61.7 & \multicolumn{1}{c}{-} & 82.6 & 61.4 & 31.1 & 83.9 & \multicolumn{1}{c}{-} & 40.8 & \multicolumn{1}{c}{-} & \textbf{38.4} & \multicolumn{1}{l|}{51.1} & \multicolumn{1}{c}{-} & 52.5 \\
\multicolumn{1}{c|}{PIT~\cite{lv2020cross}} & 83.1 & 27.6 & 81.5 & 8.9 & 0.3 & 21.8 & 26.4 & 33.8 & 76.4 & \multicolumn{1}{c}{-} & 78.8 & 64.2 & 27.6 & 79.6 & \multicolumn{1}{c}{-} & 31.2 & \multicolumn{1}{c}{-} & 31.0 & \multicolumn{1}{l|}{31.3} & 44.0 & 51.8 \\
\multicolumn{1}{c|}{\textbf{\textcolor{blue}{CBST}}~\cite{zou2018unsupervised}} & 68.0 & 29.9 & 76.3 & 10.8 & 1.4 & 33.9 & 22.8 & 29.5 & 77.6 & \multicolumn{1}{c}{-} & 78.3 & 60.6 & 28.3 & 81.6 & \multicolumn{1}{c}{-} & 23.5 & \multicolumn{1}{c}{-} & 18.8 & \multicolumn{1}{l|}{39.8} & 42.6 & 48.9 \\
\multicolumn{1}{c|}{\textbf{\textcolor{blue}{MRKLD}}~\cite{zou2019confidence}} & 67.7 & 32.2 & 73.9 & 10.7 & 1.6 & \textbf{37.4} & 22.2 & 31.2 & 80.8 & \multicolumn{1}{c}{-} & 80.5 & 60.8 & 29.1 & 82.8 & \multicolumn{1}{c}{-} & 25.0 & \multicolumn{1}{c}{-} & 19.4 & \multicolumn{1}{l|}{45.3} & 43.8 & 50.1 \\
\multicolumn{1}{c|}{\textbf{\textcolor{blue}{R-MRNet}}~\cite{zheng2020rectifying}} & 87.6 & 41.9 & 83.1 & 14.7 & 1.7 & 36.2 & 31.3 & 19.9 & 81.6 & \multicolumn{1}{c}{-} & 80.6 & 63.0 & 21.8 & 86.2 & \multicolumn{1}{c}{-} & 40.7 & \multicolumn{1}{c}{-} & 23.6 & \multicolumn{1}{l|}{53.1} & 47.9 & 54.9 \\
\multicolumn{1}{c|}{\textbf{\textcolor{blue}{DACS}}~\cite{tranheden2020dacs}} & 80.56 & 25.12 & 81.90 & 21.46 & 2.85 & 37.20 & 22.67 & 23.99 & \textbf{83.69} & \multicolumn{1}{c}{-} & \textbf{90.77} & \textbf{67.61} & \textbf{38.33} & 82.92 & \multicolumn{1}{c}{-} & 38.90 & \multicolumn{1}{c}{-} & 28.49 & \multicolumn{1}{l|}{47.58} & 48.34 & 54.81 \\
\hline \hline
\multicolumn{1}{c|}{Source Only} & 25.40 & 15.55 & 59.70 & 18.07 & 0.66 & 26.35 & 19.36 & 30.22 & 72.50 & \multicolumn{1}{c}{-} & 74.28 & 48.11 & 13.67 & 74.62 & \multicolumn{1}{c}{-} & 36.94 & \multicolumn{1}{c}{-} & 13.92 & \multicolumn{1}{l|}{36.45} & 35.36 & 40.06 \\
\multicolumn{1}{c|}{EnD~\cite{hinton2015distilling}} & 85.29 & 25.47 & 81.52 & 15.66 & 3.94 & 34.87 & 30.08 & 35.41 & 80.18 & \multicolumn{1}{c}{-} & 85.86 & 59.94 & 22.78 & 83.53 & \multicolumn{1}{c}{-} & 36.58 & \multicolumn{1}{c}{-} & 16.88 & \multicolumn{1}{l|}{52.42} & 46.90 & 53.53 \\
\multicolumn{1}{c|}{EnD$^{2}$~\cite{malinin2019ensemble}} & 83.88 & 39.08 & 81.14 & 12.65 & 1.01 & 41.16 & 22.91 & 28.26 & 82.83 & \multicolumn{1}{c}{-} & 84.17 & 69.54 & 23.87 & 87.65 & \multicolumn{1}{c}{-} & 41.59 & \multicolumn{1}{c}{-} & 21.68 & \multicolumn{1}{l|}{53.26} & 48.42 & 55.38 \\ 
\multicolumn{1}{c|}{Ours (Pixel)} & 86.98 & 44.18 & 80.95 & 19.38 & 1.52 & 30.47 & 25.64 & 30.39 & 79.92 & \multicolumn{1}{c}{-} & 78.84 & 56.55 & 27.14 & 84.47 & \multicolumn{1}{c}{-} & 44.52 & \multicolumn{1}{c}{-} & 26.03 & \multicolumn{1}{l|}{55.08} & 48.25 & 55.43 \\
\multicolumn{1}{c|}{Ours (Channel)} & \textbf{88.65} & \textbf{46.69} & \textbf{83.79} & \textbf{22.66} &  \textbf{4.14} & 35.01 & \textbf{35.93} & \textbf{36.16} & 82.80 & \multicolumn{1}{c}{-} & 81.35 & 61.61 & 32.13 & \textbf{87.93} & \multicolumn{1}{c}{-} & \textbf{52.79} & \multicolumn{1}{c}{-} & 31.95 & \multicolumn{1}{l|}{\textbf{57.65}} & \textbf{52.58} & \textbf{59.95} \\ \mytoprule
\end{tabular}}
\caption{The quantitative results evaluated on the GTA5$\to$Cityscapes and SYNTHIA$\to$Cityscapes benchmarks. The numbers presented in the middle and the last two columns correspond to per-class IoUs, mIoU, and mIoU*, respectively. mIoU* represents the mean IoU over all the semantic classes excluding those with superscript *, and is adopted by a few baseline methods in their original papers. The models used in our semantic segmentation based UDA model ensemble $\teacherset$ are highlighted in \textbf{\textcolor{blue}{blue}}. The setting `\textit{Source Only}' indicates that the student model is trained only with the source domain ground truth annotations. The evaluation results of EnD~\cite{hinton2015distilling} and EnD$^2$~\cite{malinin2019ensemble} are obtained from our self-implemented models, while those of the remaining baselines are directly obtained from their original papers.}
\label{tab:benchmark}
\end{table*} 
\section{Experimental Results}
\label{sec:experiments}
In this section, we present a number of experiments to validate the design of our framework. First, we compare our framework against a number of baselines and demonstrate its superior performance. Next, according to the experimental results, we show that the output unification operation can provide robustness against the inconsistency issue. Then, we present another experiment, in which under-performing members are added to $\teacherset$ to create performance variations, to demonstrate the robustness of our framework with $\fusion^{Channel}$ against the performance variation issue. In addition, we perform experiments to validate the effectiveness of the fusion policy selection strategy and analyze how the value of $\kappa$ in the conflict resolving mechanism can impact the performance. Finally, we explore the flexibility of our framework by adding additional teacher models in an iterative manner, and show that our framework is able to evolve with time.

\subsection{Quantitative Results on the Benchmarks}
\label{sec:experiments:benchmark}
Table~\ref{tab:benchmark} first demonstrates the quantitative results of the proposed framework against a number of baselines mentioned in Section~\ref{sec:setup} on the two benchmarks: GTA5$\to$Cityscapes and SYNTHIA$\to$Cityscapes. It is observed that the student models trained under our proposed framework with $\fusion^{Channel}$ (i.e., `Ours (Channel)') is able to outperform the previous ensemble-distillation baselines, i.e., EnD~\cite{hinton2015distilling} and EnD$^2$~\cite{malinin2019ensemble}, by a margin of 6.64\% mIoU and 6.02\% mIoU on GTA5$\to$Cityscapes, and 6.41\% mIoU and 4.57\% mIoU on SYNTHIA$\to$Cityscapes, respectively. In addition, it is also observed that the student model trained with the proposed framework with $\fusion^{Channel}$ (i.e., `Ours (Channel)') is able to outperform that with the baseline $\fusion^{Pixel}$ (i.e., `Ours (Pixel)') by a margin of 4.72\% mIoU on GTA5$\to$Cityscapes, and 4.52\% mIoU on SYNTHIA$\to$Cityscapes. 

\subsection{Robustness of the Proposed Framework}
\label{sec:experiments:robustness}
\begin{figure}[t]\includegraphics[width=\linewidth]{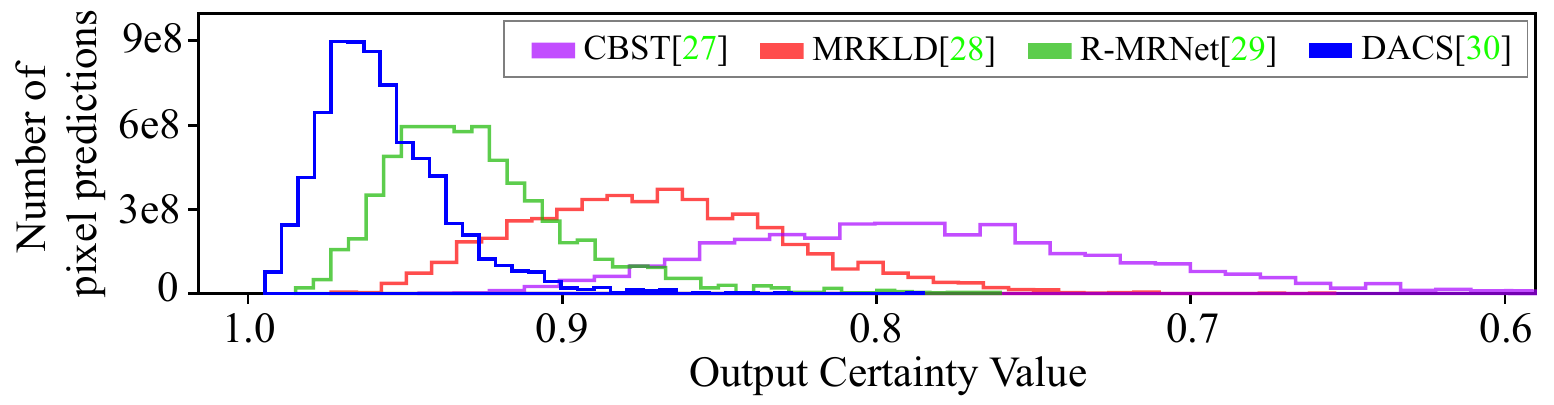}
    \caption{The distribution of the pixel output certainty values from the models in $\teacherset$, which are trained on GTA5$\to$Cityscapes, with their output certainty values normalized to the range $[0, 1]$ using softmax operation and evaluated on the training set of Cityscapes.}
    \label{fig:certainty}
\end{figure}

In this section, we validate the robustness of the proposed framework against the two issues discussed in Section~\ref{sec:introduction}. First, to verify that the proposed output unification operation can provide robustness against the inconsistency issue, we leverage the insights from an experiment conducted on $\teacherset$, with the members of $\teacherset$  bearing substantial output certainty scale variations, as shown in Fig.~\ref{fig:certainty}. The results from Table~\ref{tab:benchmark} reveal that the performance of the student models trained with `Ours~(Pixel)', which is basically the ensemble-distillation baseline EnD~\cite{hinton2015distilling} equipped with the proposed output unification method, is able to outperform EnD by a noticeable margin on both benchmarks. This implies that the adoption of the output unification method is able to provide robustness against the inconsistency issue. Nevertheless, as discussed in Section~\ref{sec:method:fusion_function}, `Ours~(Pixel)' may still be vulnerable to the performance variations of the members in $\teacherset$.

\begin{figure}[t]\includegraphics[width=\linewidth]{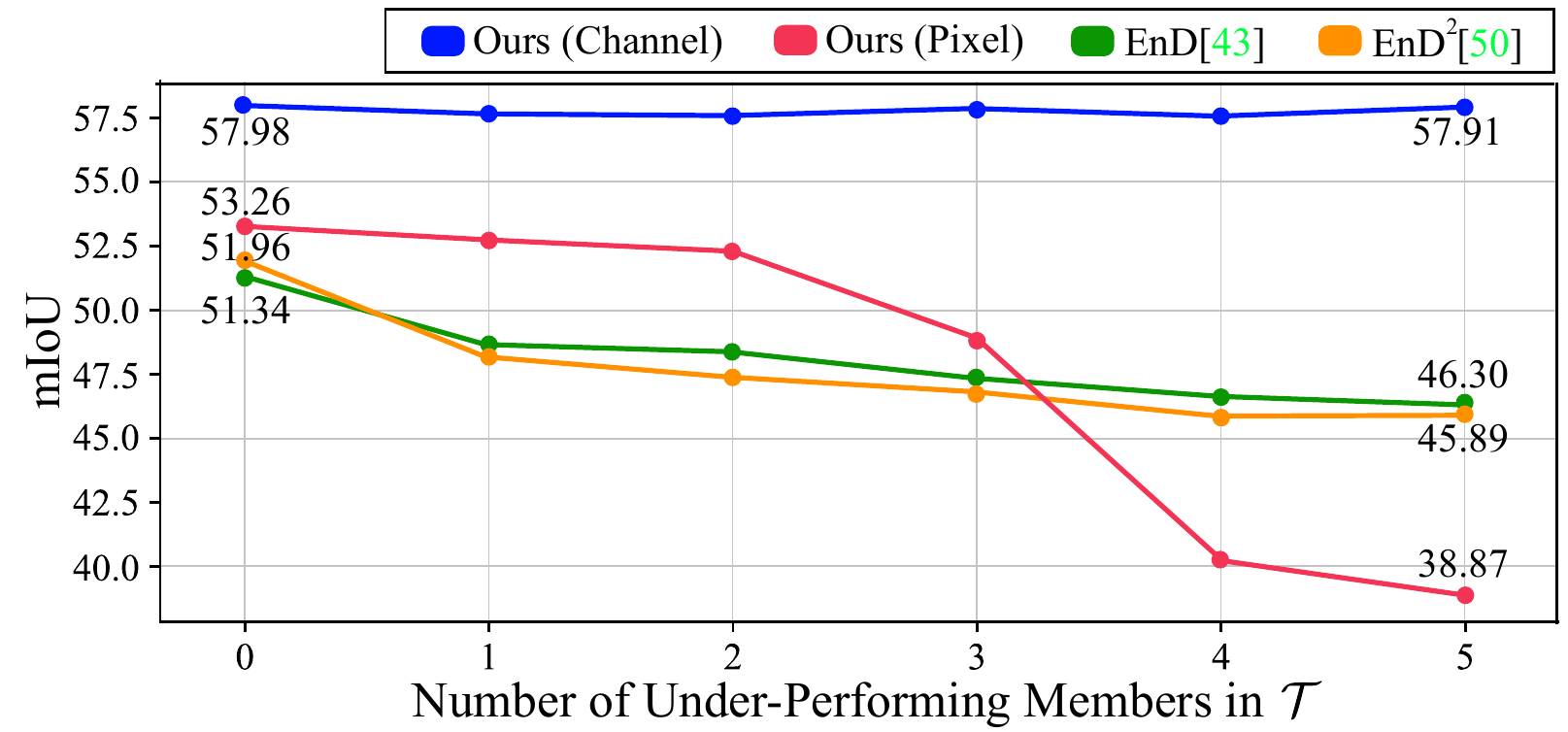}
    \caption{The performance comparison of our framework with $\fusion^{Channel}$ (`Ours~(Channel)'), our framework with $\fusion^{Pixel}$ (`Ours~(Pixel)'), and the baseline EnD and EnD$^2$ methods, with under-performing members added to $\teacherset$. In this experiment, the members in $\teacherset$ are evaluated on the GTA5$\to$Cityscapes benchmark.}
    \label{fig:robustness}
\end{figure}
\vspace{-1.5em} 
To inspect the robustness of the proposed framework with channel-wise fusion (`Ours~(Channel)') against the performance variation issue, we next look into an experiment that introduces performance variations by adding under-performing members into $\teacherset$. Specifically, we use the models trained using only source domain instances, i.e., `\textit{Source Only}' in Table~\ref{tab:benchmark}, as the under-performing members. The analysis is presented in Fig.~\ref{fig:robustness}. It is observed that the performance of the students trained with `Ours~(Pixel)', EnD, and EnD$^{2}$ all degrade when the number of under-performing members increases. In contrast, the students trained using `Ours~(Channel)' is able to maintain their performance despite the inclusion of the under-performing members in  $\teacherset$. These results demonstrate the significance of preventing clueless incorporation of information from all $t \in \teacherset$, and highlight the effectiveness of $\fusion^{Channel}$ in providing robustness against the unfavorable performance variation issue.

\subsection{Analysis on Channel-Wise Fusion}
\label{sec:experiments:analysis}
\paragraph{Fusion Policy: }~To examine the effectiveness of $\policy$ selected according to the strategy described in Section~\ref{subsubsec::certainty-aware-policy}, we design two additional fusion policies, $\policy^{rnd}$ and $\policy^{tgt}$, for comparison purposes. $\policy^{rnd}$ designates a $t \in \teacherset$ for each $c \in \classset$ randomly, while $\policy^{tgt}$ carries out this designation greedily according to the oracle performance of the models in $\teacherset$ in the target domain (i.e., $\policy^*$ described in Proposition 2). As shown in Table~\ref{tab:policy}, the mIoU of the fused pseudo labels generated based on the $\policy$ selected according to the proposed strategy is 7.09\% higher than that with $\policy^{rnd}$. It is also observed that the mIoU of the fused pseudo labels generated based on this $\policy$ is very close to the mIoU obtained from $\policy^{tgt}$. These pieces of evidence validate that the proposed policy selection strategy is able to generate a fusion policy $\policy$ that is very close to the optimal policy $\policy^*$ without leveraging any sort of unavailable target domain ground truth annotations.

\begin{table}[t]

\resizebox{\linewidth}{!}{\newcommand{\mytoprule}{\toprule[1pt]}
\renewcommand{\arraystretch}{1.05}
\centering
\footnotesize
\begin{tabular}{cccccccc}
\mytoprule
$\kappa$ & 1 & 5 & 7 & 13 & 21 & 27 \\ \hline
mIoU Gains & +0.00 & +0.15 & +0.37 & +0.73 & +0.31 & +0.09\\
\mytoprule
\end{tabular}}
\caption{The mIoU gains of $\fusion^{Channel}$ with $\policy^{rnd}$ for different $\kappa$.}
\label{tab:kernel}
\end{table}
 \begin{figure}[t]
    \includegraphics[width=\linewidth]{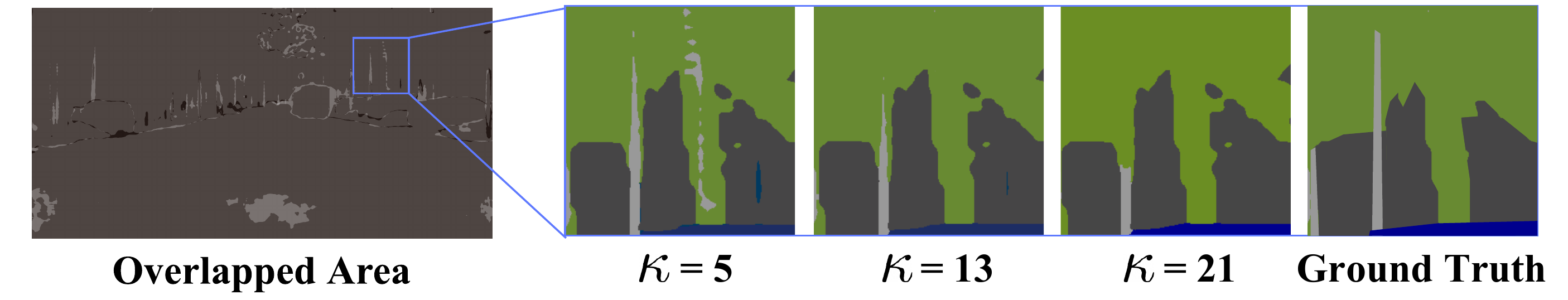}
    \caption{The visualized results of the pseudo labels generated by the proposed channel-wise fusion with different choices of $\kappa$.} 
    \label{fig:kernel}
\end{figure}
 \begin{table}[t]
\newcommand{\mytoprule}{\toprule[1.08pt]}
\renewcommand{\arraystretch}{1.1}
\centering
\footnotesize
\begin{tabular}{ccccccc}
\mytoprule
Policy & $\policy^{rnd}$ (Random) & $\policy$ (Ours) & $\policy^{tgt}$ (Oracle) \\ \hline 
mIoU & 49.22 & 56.31 & 56.48\\ \hline \hline
Overlapped Area & ${|\area{o}{\policy^{rnd}}|}$ & ${|\area{o}{\policy}|}$& ${|\area{o}{\policy^{tgt}}|}$\\ \hline
Ratio & 4.6\% & 3.7\% & 2.6\% \\
mIoU Gains & +0.73 & +1.13 & +1.21 \\\mytoprule
\end{tabular}\caption{The mIoU's of the fused pseudo labels generated by channel-wise fusion $\fusion^{Channel}$ under different fusion policies w.r.t. the ground truth annotations of the training set of Cityscapes. $\policy^{rnd}$, $\policy$, and $\policy^{tgt}$ refer to the fusion policies mentioned in Section~\ref{sec:experiments:analysis}. `Ratio' refers to the average proportion of the overlapped area in an image (i.e., $\frac{|\area{o}{\policy}|}{|\imageset|}$). The `mIoU gains' represents the mIoU gains from the adoption of the proposed conflict-resolving mechanism.}
\label{tab:policy}
\end{table}
 \paragraph{Conflict-Resolving Mechanism:}~To investigate the effectiveness of the conflict-resolving mechanism described in Section~\ref{sec:method:fusion_function} and Eq.~(\ref{eq:fused_area}), we perform parameter analysis on $\fusion^{Channel}$ with different choices of $\kappa$. In this analysis, the fusion policy adopted is $\policy^{rnd}$ to ensure that results are independent of the design of $\policy$. Fig.~\ref{fig:kernel} shows the visualized results of the fused pseudo labels with different $\kappa$. Table~\ref{tab:kernel} reports the mIoU gained for various $\kappa$ with respect to the condition $\kappa = 1$. It is observed that pixel predictions in the overlapped area can be better determined if a moderate collection of predictions from the neighboring pixels are taken into account. However, if $\kappa$ becomes too large, an excessive amount of unrelated semantic information is included in the conflict resolving mechanism, and degrades the quality of the fused pseudo labels. Table~\ref{tab:policy} shows that different policies, e.g., $\policy$ and $\policy^{tgt}$, can also benefit from the conflict-resolving mechanism, thus justifying it from a different perspective.

\begin{figure}[t]\vspace{-1.2ex}
    \includegraphics[width=\linewidth]{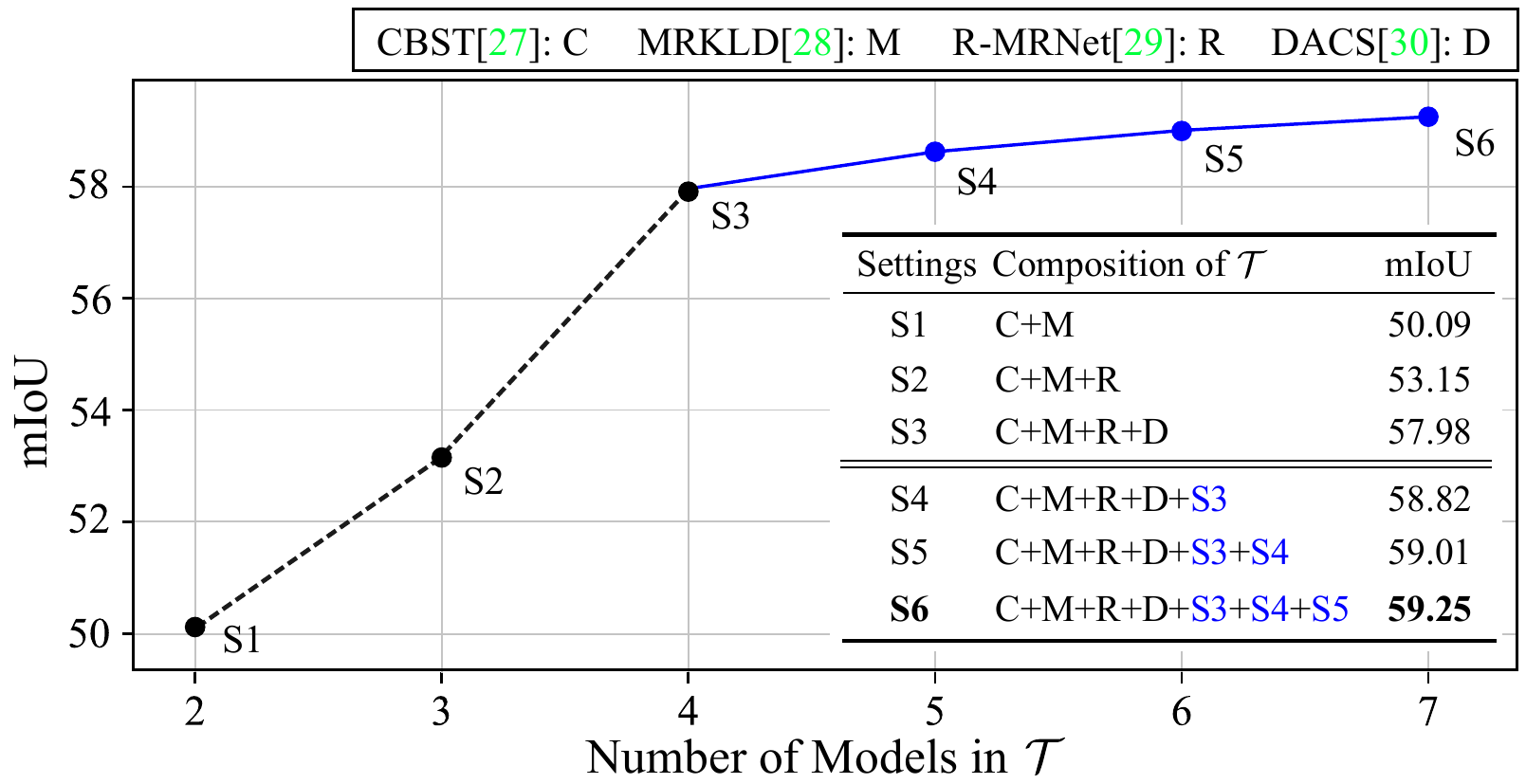}
    \caption{The performance of the student models trained with the proposed framework using $\fusion^{Channel}$ with different compositions of $\teacherset$. The experiment is performed on GTA5$\to$Cityscapes. For settings `S4', `S5' and `S6', the student models trained under the previous settings, i.e., `S3', `S4', and `S5', are added back to $\teacherset$.}
    \label{fig:flexibility}
\end{figure}
 \subsection{Flexibility of the Proposed Framework}
\label{sec:experiments:flexibility}
In comparison to end-to-end UDA ensemble learning methods, our framework is more flexible as the proposed framework can operate on any arbitrary compositions of $\teacherset$. Such flexibility enables our framework to evolve with time as (1) a model trained with any newly developed UDA method can be integrated into our framework, and (2) the student models can be added back to $\teacherset$ and further improve the overall performance. To demonstrate these two merits of our framework, we evaluate our framework with different composition of $\teacherset$, as shown in Fig.~\ref{fig:flexibility}. It is observed that the student models trained under the settings `S1', `S2', and `S3', which simulate the addition of the members trained with newly developed UDA methods, can outperform their corresponding teacher models in $\teacherset$ (whose mIoU's are reported in Table~\ref{tab:benchmark}). This implies that our framework offers the potential to evolve with time. In addition, `S4', `S5', and `S6' in Fig.~\ref{fig:flexibility} showcase that the performance of our framework can be further improved if the students are added back to $\teacherset$, indicating that our framework can be applied in an iterative manner to produce better results. These results highlight the flexibility of our framework as any UDA methods can be incorporated and potentially enhance its performance.  \section{Conclusion}
\label{sec:conclusion}
In this paper, we presented a flexible ensemble-distillation framework to address the common pitfalls, i.e., the lack of robustness, of previous methods. We incorporated an output unification operation into the proposed framework to ensure that the fused outputs of the ensemble are free of the influence from the certainty inconsistency among the models in the ensemble. In addition, to tackle the performance variation issue, we proposed a channel-wise fusion function that is robust against this issue. As our framework is able to integrate different types of UDA methods while maintaining its robustness, it therefore pioneers a new direction for future semantic segmentation based UDA researches.

\paragraph{Acknowledgement: }~The authors acknowledge the support from the Ministry of Science and Technology (MOST) in Taiwan under grant nos. MOST 110-2636-E-007-010 and MOST 110-2634-F-007-019, as well as the support from MediaTek Inc., Taiwan. The authors thank the donation of the GPUs from NVIDIA Corporation and NVIDIA AI Technology Center, and the National Center for High-performance Computing of National Applied Research Laboratories in Taiwan for providing computational and storage resources.   \clearpage

{\small
\bibliographystyle{unsrtnat}
\bibliography{egbib}

\begin{thebibliography}{55}
\providecommand{\natexlab}[1]{#1}
\providecommand{\url}[1]{\texttt{#1}}
\expandafter\ifx\csname urlstyle\endcsname\relax
  \providecommand{\doi}[1]{doi: #1}\else
  \providecommand{\doi}{doi: \begingroup \urlstyle{rm}\Url}\fi

\bibitem[Yu et~al.(2017)Yu, Koltun, and Funkhouser]{yu2017dilated}
F.~Yu, V.~Koltun, and T.~Funkhouser.
\newblock Dilated residual networks.
\newblock In \emph{Proc. IEEE Conf. Computer Vision and Pattern Recognition
  (CVPR)}, pages 472--480, Jul. 2017.

\bibitem[Lin et~al.(2017)Lin, Milan, Shen, and Reid]{lin2017refinenet}
G.~Lin, A.~Milan, C.~Shen, and I.~Reid.
\newblock Refinenet: Multi-path refinement networks for high-resolution
  semantic segmentation.
\newblock In \emph{Proc. IEEE Conf. Computer Vision and Pattern Recognition
  (CVPR)}, pages 1925--1934, Jul. 2017.

\bibitem[Yu and Koltun(2016)]{yu2015multi}
F.~Yu and V.~Koltun.
\newblock Multi-scale context aggregation by dilated convolutions.
\newblock \emph{Proc. Int. Conf. Learning Representations (ICLR)}, May 2016.

\bibitem[Badrinarayanan et~al.(2017)Badrinarayanan, Kendall, and
  Cipolla]{badrinarayanan2017segnet}
V.~Badrinarayanan, A.~Kendall, and R.~Cipolla.
\newblock Segnet: A deep convolutional encoder-decoder architecture for image
  segmentation.
\newblock \emph{IEEE Trans. Pattern Analysis and Machine Intelligence (TPAMI)},
  39\penalty0 (12):\penalty0 2481--2495, Jan. 2017.

\bibitem[Long et~al.(2015)Long, Shelhamer, and Darrell]{long2015fully}
J.~Long, E.~Shelhamer, and T.~Darrell.
\newblock Fully convolutional networks for semantic segmentation.
\newblock In \emph{Proc. IEEE Conf. Computer Vision and Pattern Recognition
  (CVPR)}, pages 3431--3440, Jun. 2015.

\bibitem[Yuan et~al.(2020)Yuan, Chen, and Wang]{yuan2019object}
Y.~Yuan, X.~Chen, and J.~Wang.
\newblock Object-contextual representations for semantic segmentation.
\newblock In \emph{Proc. European Conf. Computer Vision (ECCV)}, pages
  173--190, Oct. 2020.

\bibitem[Wu et~al.(2019)Wu, Shen, and Van Den~Hengel]{wu2019wider}
Z.~Wu, C.~Shen, and A.~Van Den~Hengel.
\newblock Wider or deeper: Revisiting the resnet model for visual recognition.
\newblock \emph{Pattern Recognition}, 90:\penalty0 119--133, 2019.

\bibitem[Sandler et~al.(2018)Sandler, Howard, Zhu, Zhmoginov, and
  Chen]{sandler2018mobilenetv2}
M.~Sandler, A.~Howard, M.~Zhu, A.~Zhmoginov, and L.-C. Chen.
\newblock Mobilenetv2: Inverted residuals and linear bottlenecks.
\newblock In \emph{Proc. IEEE Conf. Computer Vision and Pattern Recognition
  (CVPR)}, pages 4510--4520, Jun. 2018.

\bibitem[Chen et~al.(2014)Chen, Papandreou, Kokkinos, Murphy, and
  Yuille]{chen2014semantic}
L.-C. Chen, G.~Papandreou, I.~Kokkinos, K.~Murphy, and A.~L. Yuille.
\newblock Semantic image segmentation with deep convolutional nets and fully
  connected {CRFs}.
\newblock \emph{arXiv:1412.7062}, Jul. 2014.

\bibitem[Chen et~al.(2017{\natexlab{a}})Chen, Papandreou, Kokkinos, Murphy, and
  Yuille]{chen2017deeplab}
L.-C. Chen, G.~Papandreou, I.~Kokkinos, K.~Murphy, and A.~L. Yuille.
\newblock Deeplab: Semantic image segmentation with deep convolutional nets,
  atrous convolution, and fully connected {CRFs}.
\newblock \emph{IEEE Trans. Pattern Analysis and Machine Intelligence (TPAMI)},
  40\penalty0 (4):\penalty0 834--848, Apr. 2017{\natexlab{a}}.

\bibitem[Chen et~al.(2017{\natexlab{b}})Chen, Papandreou, Schroff, and
  Adam]{chen2017rethinking}
L.-C. Chen, G.~Papandreou, F.~Schroff, and H.~Adam.
\newblock Rethinking atrous convolution for semantic image segmentation.
\newblock \emph{arXiv:1706.05587}, Dec. 2017{\natexlab{b}}.

\bibitem[Chen et~al.(2018)Chen, Zhu, Papandreou, Schroff, and
  Adam]{chen2018encoder}
L.-C. Chen, Y.~Zhu, G.~Papandreou, F.~Schroff, and H.~Adam.
\newblock Encoder-decoder with atrous separable convolution for semantic image
  segmentation.
\newblock In \emph{Proc. European Conf. Computer Vision (ECCV)}, pages
  801--818, Sep. 2018.

\bibitem[Zhao et~al.(2017)Zhao, Shi, Qi, Wang, and Jia]{zhao2017pyramid}
H.~Zhao, J.~Shi, X.~Qi, X.~Wang, and J.~Jia.
\newblock Pyramid scene parsing network.
\newblock In \emph{Proc. IEEE Conf. Computer Vision and Pattern Recognition
  (CVPR)}, pages 2881--2890, Jul. 2017.

\bibitem[Hoffman et~al.(2016)Hoffman, Wang, Yu, and Darrell]{hoffman2016fcns}
J.~Hoffman, D.~Wang, F.~Yu, and T.~Darrell.
\newblock {FCNs} in the wild: Pixel-level adversarial and constraint-based
  adaptation.
\newblock \emph{arXiv:1612.02649}, Dec. 2016.

\bibitem[Hoffman et~al.(2018)Hoffman, Tzeng, Park, Zhu, Isola, Saenko, Efros,
  and Darrell]{hoffman2018cycada}
J.~Hoffman, E.~Tzeng, T.~Park, J.-Y. Zhu, P.~Isola, K.~Saenko, A.~Efros, and
  T.~Darrell.
\newblock {CyCADA}: Cycle-consistent adversarial domain adaptation.
\newblock In \emph{Proc. Int. Conf. Machine Learning (ICML)}, pages 1989--1998,
  Jul. 2018.

\bibitem[Luo et~al.(2019{\natexlab{a}})Luo, Liu, Guan, Yu, and
  Yang]{luo2019significance}
Y.~Luo, P.~Liu, T.~Guan, J.~Yu, and Y.~Yang.
\newblock Significance-aware information bottleneck for domain adaptive
  semantic segmentation.
\newblock In \emph{Proc. IEEE Int. Conf. Computer Vision (ICCV)}, pages
  6778--6787, Oct. 2019{\natexlab{a}}.

\bibitem[Gong et~al.(2019)Gong, Li, Chen, and Gool]{gong2019dlow}
R.~Gong, W.~Li, Y.~Chen, and L.~V. Gool.
\newblock Dlow: Domain flow for adaptation and generalization.
\newblock In \emph{Proc. IEEE Conf. Computer Vision and Pattern Recognition
  (CVPR)}, pages 2477--2486, Jun. 2019.

\bibitem[Wu et~al.(2018)Wu, Han, Lin, Gokhan~Uzunbas, Goldstein, Nam~Lim, and
  Davis]{wu2018dcan}
Z.~Wu, X.~Han, Y.-L. Lin, M.~Gokhan~Uzunbas, T.~Goldstein, S.~Nam~Lim, and
  L.~S. Davis.
\newblock Dcan: Dual channel-wise alignment networks for unsupervised scene
  adaptation.
\newblock In \emph{Proc. European Conf. Computer Vision (ECCV)}, pages
  518--534, Oct. 2018.

\bibitem[Tsai et~al.(2018)Tsai, Hung, Schulter, Sohn, Yang, and
  Chandraker]{tsai2018learning}
Yi-Hsuan Tsai, Wei-Chih Hung, Samuel Schulter, Kihyuk Sohn, Ming-Hsuan Yang,
  and Manmohan Chandraker.
\newblock Learning to adapt structured output space for semantic segmentation.
\newblock In \emph{Proc. of the IEEE Conf. on Computer Vision and Pattern
  Recognition (CVPR)}, pages 7472--7481, 2018.

\bibitem[Yang et~al.(2020)Yang, Xu, Li, Qi, Shen, Li, and
  Lin]{yang2020adversarial}
J.~Yang, R.~Xu, R.~Li, X.~Qi, X.~Shen, G.~Li, and L.~Lin.
\newblock An adversarial perturbation oriented domain adaptation approach for
  semantic segmentation.
\newblock In \emph{Proc. the Thirty-Fourth {AAAI} Conf. Artificial Intelligence
  (AAAI)}, pages 12613--12620, Feb. 2020.

\bibitem[Tsai et~al.(2019)Tsai, Sohn, Schulter, and Chandraker]{tsai2019domain}
Y.-H. Tsai, K.~Sohn, S.~Schulter, and M.~Chandraker.
\newblock Domain adaptation for structured output via discriminative patch
  representations.
\newblock In \emph{Proc. IEEE Int. Conf. Computer Vision (ICCV)}, pages
  1456--1465, Oct. 2019.

\bibitem[Vu et~al.(2019{\natexlab{a}})Vu, Jain, Bucher, Cord, and
  P{\'e}rez]{vu2019advent}
T.-H. Vu, H.~Jain, M.~Bucher, M.~Cord, and P.~P{\'e}rez.
\newblock Advent: Adversarial entropy minimization for domain adaptation in
  semantic segmentation.
\newblock In \emph{Proc. IEEE Conf. Computer Vision and Pattern Recognition
  (CVPR)}, pages 2517--2526, Jun. 2019{\natexlab{a}}.

\bibitem[Zhang et~al.(2018)Zhang, Qiu, Yao, Liu, and Mei]{zhang2018fully}
Y.~Zhang, Z.~Qiu, T.~Yao, D.~Liu, and T.~Mei.
\newblock Fully convolutional adaptation networks for semantic segmentation.
\newblock In \emph{Proc. IEEE Conf. Computer Vision and Pattern Recognition
  (CVPR)}, pages 6810--6818, Jun. 2018.

\bibitem[Chen et~al.(2017{\natexlab{c}})Chen, Chen, Chen, Tsai, Frank~Wang, and
  Sun]{chen2017no}
Y.-H. Chen, W.-Y. Chen, Y.-T. Chen, B.-C. Tsai, Y.-C. Frank~Wang, and M.~Sun.
\newblock No more discrimination: Cross city adaptation of road scene
  segmenters.
\newblock In \emph{Proc. IEEE Int. Conf. Computer Vision (ICCV)}, pages
  1992--2001, Oct. 2017{\natexlab{c}}.

\bibitem[Zheng and Yang(2020{\natexlab{a}})]{zheng2019unsupervised}
Z.~Zheng and Y.~Yang.
\newblock Unsupervised scene adaptation with memory regularization in vivo.
\newblock \emph{Proc. Int. Joint Conf. Artificial Intelligence (IJCAI)}, Jul.
  2020{\natexlab{a}}.

\bibitem[Choi et~al.(2019)Choi, Kim, and Kim]{choi2019self}
J.~Choi, T.~Kim, and C.~Kim.
\newblock Self-ensembling with gan-based data augmentation for domain
  adaptation in semantic segmentation.
\newblock In \emph{Proc. IEEE Int. Conf. Computer Vision (ICCV)}, pages
  6830--6840, Oct. 2019.

\bibitem[Zou et~al.(2018)Zou, Yu, Vijaya~Kumar, and Wang]{zou2018unsupervised}
Y.~Zou, Z.~Yu, B.~Vijaya~Kumar, and J.~Wang.
\newblock Unsupervised domain adaptation for semantic segmentation via
  class-balanced self-training.
\newblock In \emph{Proc. European Conf. Computer Vision (ECCV)}, pages
  289--305, Sep. 2018.

\bibitem[Zou et~al.(2019)Zou, Yu, Liu, Kumar, and Wang]{zou2019confidence}
Y.~Zou, Z.~Yu, X.~Liu, B.~Kumar, and J.~Wang.
\newblock Confidence regularized self-training.
\newblock In \emph{Proc. of the IEEE Int. Conf. on Computer Vision (ICCV)},
  pages 5982--5991, Oct. 2019.

\bibitem[Zheng and Yang(2020{\natexlab{b}})]{zheng2020rectifying}
Z.~Zheng and Y.~Yang.
\newblock Rectifying pseudo label learning via uncertainty estimation for
  domain adaptive semantic segmentation.
\newblock \emph{Int. Journal of Computer Vision (IJCV)}, 2020{\natexlab{b}}.
\newblock \doi{10.1007/s11263-020-01395-y}.

\bibitem[Tranheden et~al.(2021)Tranheden, Olsson, Pinto, and
  Svensson]{tranheden2020dacs}
W.~Tranheden, V.~Olsson, J.~Pinto, and L.~Svensson.
\newblock {DACS}: Domain adaptation via cross-domain mixed sampling.
\newblock In \emph{Proc. IEEE Winter Conf. on Applications of Computer Vision
  (WACV)}, pages 1379--1389, Jan. 2021.

\bibitem[Luo et~al.(2019{\natexlab{b}})Luo, Zheng, Guan, Yu, and
  Yang]{luo2019taking}
Y.~Luo, L.~Zheng, T.~Guan, J.~Yu, and Y.~Yang.
\newblock Taking a closer look at domain shift: Category-level adversaries for
  semantics consistent domain adaptation.
\newblock In \emph{Proc. of the IEEE Conf. on Computer Vision and Pattern
  Recognition (CVPR)}, pages 2507--2516, 2019{\natexlab{b}}.

\bibitem[Lee et~al.(2018)Lee, Ros, Li, and Gaidon]{lee2018spigan}
K.-H. Lee, G.~Ros, J.~Li, and A.~Gaidon.
\newblock Spigan: Privileged adversarial learning from simulation.
\newblock \emph{arXiv:1810.03756}, 2018.

\bibitem[Chen et~al.(2019{\natexlab{a}})Chen, Li, Chen, and
  Gool]{chen2019learning}
Y.~Chen, W.~Li, X.~Chen, and L.~V. Gool.
\newblock Learning semantic segmentation from synthetic data: A geometrically
  guided input-output adaptation approach.
\newblock In \emph{Proc. IEEE Conf. on Computer Vision and Pattern Recognition
  (CVPR)}, pages 1841--1850, 2019{\natexlab{a}}.

\bibitem[Vu et~al.(2019{\natexlab{b}})Vu, Jain, Bucher, Cord, and
  P{\'e}rez]{vu2019dada}
T.-H. Vu, H.~Jain, M.~Bucher, M.~Cord, and P.~P{\'e}rez.
\newblock Dada: Depth-aware domain adaptation in semantic segmentation.
\newblock In \emph{Proc. IEEE Int. Conf. on Computer Vision (ICCV)}, pages
  7364--7373, 2019{\natexlab{b}}.

\bibitem[Lv et~al.(2020)Lv, Liang, Chen, and Lin]{lv2020cross}
F.~Lv, T.~Liang, X.~Chen, and G.~Lin.
\newblock Cross-domain semantic segmentation via domain-invariant interactive
  relation transfer.
\newblock In \emph{Proc. of the IEEE Conf. on Computer Vision and Pattern
  Recognition (CVPR)}, pages 4334--4343, 2020.

\bibitem[Yang and Soatto(2020)]{yang2020fda}
Y.~Yang and S.~Soatto.
\newblock {FDA}: Fourier domain adaptation for semantic segmentation.
\newblock In \emph{Proc. of the IEEE Conf. on Computer Vision and Pattern
  Recognition (CVPR)}, pages 4085--4095, 2020.

\bibitem[Chen et~al.(2019{\natexlab{b}})Chen, Lin, Yang, and
  Huang]{chen2019crdoco}
Y.-C. Chen, Y.-Y. Lin, M.-H. Yang, and J.-B. Huang.
\newblock Crdoco: Pixel-level domain transfer with cross-domain consistency.
\newblock In \emph{Proc. IEEE Conf. Computer Vision and Pattern Recognition
  (CVPR)}, pages 1791--1800, Jun. 2019{\natexlab{b}}.

\bibitem[Li et~al.(2019)Li, Yuan, and Vasconcelos]{li2019bidirectional}
Y.~Li, L.~Yuan, and N.~Vasconcelos.
\newblock Bidirectional learning for domain adaptation of semantic
  segmentation.
\newblock In \emph{Proc. IEEE Conf. Computer Vision and Pattern Recognition
  (CVPR)}, pages 6936--6945, Jun. 2019.

\bibitem[Du et~al.(2019)Du, Tan, Yang, Feng, Xue, Zheng, Ye, and
  Zhang]{du2019ssf}
L.~Du, J.~Tan, H.~Yang, J.~Feng, X.~Xue, Q.~Zheng, X.~Ye, and X.~Zhang.
\newblock Ssf-dan: Separated semantic feature based domain adaptation network
  for semantic segmentation.
\newblock In \emph{Proc. IEEE Int. Conf. Computer Vision (ICCV)}, pages
  982--991, Oct. 2019.

\bibitem[Nguyen-Meidine et~al.(2021)Nguyen-Meidine, Belal, Kiran, Dolz,
  Blais-Morin, and Granger]{nguyen2021unsupervised}
L.~T. Nguyen-Meidine, A.~Belal, M.~Kiran, J.~Dolz, L.-A. Blais-Morin, and
  E.~Granger.
\newblock Unsupervised multi-target domain adaptation through knowledge
  distillation.
\newblock In \emph{Proc. IEEE Winter Conf. on Applications of Computer Vision
  (WACV)}, pages 1339--1347, 2021.

\bibitem[Kang et~al.(2019)Kang, Jiang, Yang, and
  Hauptmann]{kang2019contrastive}
G.~Kang, L.~Jiang, Y.~Yang, and A.~G Hauptmann.
\newblock Contrastive adaptation network for unsupervised domain adaptation.
\newblock In \emph{Proc. the IEEE Conf. on Computer Vision and Pattern
  Recognition (CVPR)}, pages 4893--4902, 2019.

\bibitem[Bucilu\v{a} et~al.(2006)Bucilu\v{a}, Caruana, and
  Niculescu-Mizil]{bucilua2006model}
C.~Bucilu\v{a}, R.~Caruana, and A.~Niculescu-Mizil.
\newblock Model compression.
\newblock In \emph{Proc. 12th ACM SIGKDD Int. Conf. Knowledge Discovery and
  Data mining (KDD-06)}, pages 535--541, Aug. 2006.

\bibitem[Hinton et~al.(2015)Hinton, Vinyals, and Dean]{hinton2015distilling}
G.~Hinton, O.~Vinyals, and J.~Dean.
\newblock Distilling the knowledge in a neural network.
\newblock \emph{arXiv:1503.02531}, Mar. 2015.

\bibitem[Cho and Hariharan(2019)]{cho2019efficacy}
J.~H. Cho and B.~Hariharan.
\newblock On the efficacy of knowledge distillation.
\newblock In \emph{Proc. IEEE Int. Conf. Computer Vision (ICCV)}, pages
  4794--4802, Oct. 2019.

\bibitem[Furlanello et~al.(2018)Furlanello, Lipton, Tschannen, Itti, and
  Anandkumar]{furlanello2018born}
T.~Furlanello, Z.~C. Lipton, M.~Tschannen, L.~Itti, and A.~Anandkumar.
\newblock Born again neural networks.
\newblock \emph{Proc. Int. Conf. Machine Learning (ICML)}, Jul. 2018.

\bibitem[Balan et~al.(2015)Balan, Rathod, Murphy, and
  Welling]{balan2015bayesian}
A.~K. Balan, V.~Rathod, K.~P. Murphy, and M.~Welling.
\newblock Bayesian dark knowledge.
\newblock In \emph{Proc. Advances in Neural Information Processing Systems
  (NeurIPS)}, pages 3438--3446, Dec. 2015.

\bibitem[Nguyen-Meidine et~al.(2020)Nguyen-Meidine, Granger, Kiran, Dolz, and
  Blais-Morin]{nguyen2020joint}
Le~Thanh Nguyen-Meidine, {\'E}ric Granger, M.~Kiran, J.~Dolz, and Louis-Antoine
  Blais-Morin.
\newblock Joint progressive knowledge distillation and unsupervised domain
  adaptation.
\newblock \emph{Int. Joint Conf. on Neural Networks (IJCNN)}, pages 1--8, 2020.

\bibitem[Orbes-Arteainst et~al.(2019)Orbes-Arteainst, Cardoso, S{\o}rensen,
  Igel, Ourselin, Modat, Nielsen, and Pai]{orbes2019knowledge}
M.~Orbes-Arteainst, J.~Cardoso, L.~S{\o}rensen, C.~Igel, S.~Ourselin, M.~Modat,
  M.~Nielsen, and A.~Pai.
\newblock Knowledge distillation for semi-supervised domain adaptation.
\newblock In \emph{OR 2.0 Context-Aware Operating Theaters and Machine Learning
  in Clinical Neuroimaging}, pages 68--76, 2019.

\bibitem[Liu et~al.(2019)Liu, Chen, Liu, Qin, Luo, and Wang]{liu2019structured}
Y.~Liu, K.~Chen, C.~Liu, Z.~Qin, Z.~Luo, and J.~Wang.
\newblock Structured knowledge distillation for semantic segmentation.
\newblock In \emph{Proc. IEEE Conf. Computer Vision and Pattern Recognition
  (CVPR)}, pages 2604--2613, Jun. 2019.

\bibitem[Malinin et~al.(2020)Malinin, Mlodozeniec, and
  Gales]{malinin2019ensemble}
A.~Malinin, B.~Mlodozeniec, and M.~Gales.
\newblock Ensemble distribution distillation.
\newblock In \emph{Proc. Int. Conf. Learning Representations (ICLR)}, 2020.

\bibitem[Richter et~al.(2016)Richter, Vineet, Roth, and
  Koltun]{richter2016playing}
S.~R. Richter, V.~Vineet, S.~Roth, and V.~Koltun.
\newblock Playing for data: Ground truth from computer games.
\newblock In \emph{Proc. European Conf. Computer Vision (ECCV)}, pages
  102--118, Oct. 2016.

\bibitem[Cordts et~al.(2016)Cordts, Omran, Ramos, Rehfeld, Enzweiler, Benenson,
  Franke, Roth, and Schiele]{Cordts2016Cityscapes}
M.~Cordts, M.~Omran, S.~Ramos, T.~Rehfeld, M.~Enzweiler, R.~Benenson,
  U.~Franke, S.~Roth, and B.~Schiele.
\newblock The cityscapes dataset for semantic urban scene understanding.
\newblock In \emph{Proc. IEEE Conf. Computer Vision and Pattern Recognition
  (CVPR)}, Jun. 2016.

\bibitem[Ros et~al.(2016)Ros, Sellart, Materzynska, Vazquez, and
  Lopez]{ros2016synthia}
G.~Ros, L.~Sellart, J.~Materzynska, D.~Vazquez, and A.~M. Lopez.
\newblock {The SYNTHIA Dataset}: A large collection of synthetic images for
  semantic segmentation of urban scenes.
\newblock In \emph{Proc. IEEE Conf. Computer Vision and Pattern Recognition
  (CVPR)}, pages 3234--3243, Jun. 2016.

\bibitem[Lee(2013)]{lee2013pseudo}
D.-H. Lee.
\newblock Pseudo-label: The simple and efficient semi-supervised learning
  method for deep neural networks.
\newblock In \emph{Workshop on challenges in representation learning, Int.
  Conf. on Machine Learning (ICML)}, volume~3, Jun. 2013.

\bibitem[Malinin and Gales(2018)]{malinin2018predictive}
A.~Malinin and M.~Gales.
\newblock Predictive uncertainty estimation via prior networks.
\newblock In \emph{Proc. Advances in Neural Information Processing Systems
  (NeurIPS)}, pages 7047--7058, 2018.

\end{thebibliography}
}

\clearpage



\section*{Supplementary material}

# Supplementary Material

# Table of Contents

| Symbol | Definition |
| --- | --- |
| $c$ | A class. |
| $\mathcal{C}$ | The set of all $c$. |
| $t$ | A teacher model. |
| $\mathcal{T}$ | The set of all $t$. |
| $p$ | A pixel in an image. |
| $\mathcal{I}$ | The set of all $p$. |
| $x_{src}$ | Source domain image. |
| $y_{src}$ | Source domain ground truth. |
| $x_{tgt}$ | Target domain image. |
| $y_{tgt}$ | Target domain ground truth. |
| $\mathcal{D}_{src}$ | Source domain dataset. |
| $\mathcal{D}_{tgt}$ | Target domain dataset. |
| $f^{Pixel}$ | Pixel-wise fusion. |
| $f^{Channel}$ | Channel-wise fusion. |
| $\pi$ | A fusion policy. |
| $A$ | A segmentation map. |
| $A_c$ | The segmentation map of class $c$. |
| $A_c^{gt}$ | The ground truth of $A_c$. |
| $A_c^\pi$ | $A_c^\pi := \{p \,|\, p \in \mathcal{I}, \pi(c) = t, \hat{y}^{(p,c,t)} = 1\}$ is the pseudo label of class $c$ selected according to policy $\pi$. |
| $A_o^\pi$ | $A_o^\pi := \bigcup_{\substack{c_1 \neq c_2, \\ c_1, c_2 \in \mathcal{C}}} (A_{c_1}^\pi \cap A_{c_2}^\pi)$ is the overlapped area between $A_c^\pi$. |
| $A_{o,c}^\pi$ | $A_{o,c}^\pi := A_o^\pi \cap A_c^\pi$ is the overlapped area of a class $c$. |
| $\Phi$ | The function that calculates the IoU of a segmentation map with its ground truth annotation. |
| $\tilde{\Phi}$ | The IoU of the fused results generated using $f^{Channel}$ w.r.t. $y_{tgt}$. |
| $c_0$ | An unlabeled symbol. |
| $\varepsilon$ | A class label to be assigned in $A_o^\pi$ under the formulation of $f^{Channel}$. |
| $\mathcal{C}_p^\pi$ | $\mathcal{C}_p^\pi := \{c \,|\, c \in \mathcal{C}; p \in A_c^\pi\}$ is the set of class(es) that collects the class label(s) in $p \in A_c^\pi$. |

Table S1: List of commonly-used symbols.

# S1 Background Material

In this section, we walk through the background material of the previous semantic segmentation based unsupervised domain adaptation (UDA) methods. We first offer an overview of the concepts behind the adversarial domain adaptation (ADA) methods in Section S1.1. Then, we review the pseudo labeling strategy in Section S1.2. Some commonly used symbols are summarized in Table S1.

## S1.1 The Concepts Behind the Adversarial Domain Adaptation Methods

For the semantic segmentation based UDA problem considered in this paper, the models are granted accesses to the image-label pairs $x_{src} \in \mathbb{R}^{|\mathcal{I}| \times 3}, y_{src} \in \{0, 1\}^{|\mathcal{I}| \times |\mathcal{C}|}$ from a source domain dataset $\mathcal{D}_{src}$, and the images $x_{tgt} \in \mathbb{R}^{|\mathcal{I}| \times 3}$ from a target domain dataset $\mathcal{D}_{tgt}$, where $\mathcal{I}$ is the set of pixels in an image, and $\mathcal{C}$ is a given set of semantic classes. The goal is to train a model $G_\theta$ parameterized by $\theta$, from which the semantic segmentation predictions can best estimate the target domain ground truth $y_{tgt}$. For example, in AdaptSegNet [1], a generator $G_\theta : \mathbb{R}^{|\mathcal{I}| \times 3} \to \mathbb{R}^{|\mathcal{I}| \times |\mathcal{C}|}$ is trained against a discriminator $D_\theta : \mathbb{R}^{|\mathcal{I}| \times |\mathcal{C}|} \to \mathbb{R}^2$ using an adversarial training scheme for minimizing the domain gap. The training objective of $D_\theta$ is to distinguish whether the semantic segmentation outputs from $G_\theta$ belong to the source domain or not. In contrast, the training objectives of $G_\theta$ is to confuse the discriminator $D_\theta$ with its predictions. Their loss functions $L_G, L_D$ are defined as follows, respectively:

$$L_G = -\sum_{p \in \mathcal{I}} \log D^0(\hat{s}_{tgt}^{(p,c)}), \tag{S1}$$

$$L_D = -\sum_{p \in \mathcal{I}} (1-z) \log D^0(\hat{s}_{tgt}^{(p,c)}) + z \log D^1(\hat{s}_{src}^{(p,c)}), \tag{S2}$$

where $c \in \mathcal{C}$ denotes a class, $p \in \mathcal{I}$ denotes a pixel in an image, $\hat{s}_{src}^{(p,c)} = G_\theta(x_{src}) \in \mathbb{R}^{|\mathcal{I}| \times |\mathcal{C}|}$ is the softmax output of $G_\theta$ for $x_{src}$, and $\hat{s}_{tgt}^{(p,c)} = G_\theta(x_{tgt}) \in \mathbb{R}^{|\mathcal{I}| \times |\mathcal{C}|}$ is the softmax output of $G_\theta$ for $x_{tgt}$. $D^0, D^1$ denote the first and second output channels of $D_\theta$, which represent the certainty of $D_\theta$ on whether the input is drawn from $\mathcal{D}_{tgt}$ or $\mathcal{D}_{src}$, respectively. The binary indicator $z$ is either zero or one to indicate that the samples are drawn from the target or the source domains, respectively.

## S1.2 Pseudo Labeling Strategy

Pseudo labeling was pioneered in [2] for improving the performance of classification tasks. For the semantic segmentation based UDA problem, pseudo labeling is a common measure used in the fine-tuning phase by several self-training methods. During the fine-tuning phase, a model is trained to minimize the loss between the pseudo labels ($\hat{y}_{tgt}$) and the predictions of the model on target domain instances ($x_{tgt}$). These pseudo labels are generated by taking the $\arg\max$ operation over the softmax predictions $\hat{s}_{tgt}$ of the model, which can be formulated as the following equation:

$$\hat{y}_{tgt}^{(p,c)} = \begin{cases} 1, & \text{if } c = \arg\max_{c \in \mathcal{C}} \{\hat{s}_{tgt}^{(p,c)}\} \\ 0, & \text{otherwise} \end{cases}, \tag{S3}$$

where $\hat{s}_{tgt}^{(p,c)} = m_\theta(x_{tgt}) \in \mathbb{R}^{|\mathcal{I}| \times |\mathcal{C}|}$ is the softmax output from a segmentation model $m_\theta : \mathbb{R}^{|\mathcal{I}| \times 3} \to \mathbb{R}^{|\mathcal{I}| \times |\mathcal{C}|}$, which is the model parameterized by $\theta$ to be fine-tuned in the target domain. By reducing the cross-entropy loss between the predictions $\hat{s}_{tgt}$ and the one-hot pseudo labels $\hat{y}_{tgt}$, the decision boundaries of the model $m_\theta$ are adjusted to lie in low-density regions [2]. This additional fine-tuning stage encourages the model $m_\theta$ to produce high-certainty predictions, and enhances its stability in deployment time.

## S2 Theoretical Properties of Channel-Wise Fusion

In this section, we provide detailed descriptions of the theoretical properties of the proposed channel-wise fusion function (i.e., $f^{Channel}$). We first define the evaluation metric for semantic segmentation maps, i.e., mIoU, in Section S2.1. Next, we elaborate on the differences between $f^{Channel}$ and $f^{Pixel}$ in Section S2.2. Then, we discuss how the conflict-resolving mechanism can influence the effectiveness of $f^{Channel}$ in Section S2.3. Finally, in Section S2.4, we investigate the properties of the proposed $f^{Channel}$ under the condition that $|A_o^\pi| = 0$, and derive the proofs for **Proposition 1** and **Proposition 2** mentioned in the main manuscript.

### S2.1 Mean Intersection Over Union

In this section, we provide the definition and detailed explanation of the commonly used evaluation metric *mIoU* for semantic segmentation maps. Given a segmentation map $A \in 2^\mathcal{I} \times \mathcal{C}$ with $|\mathcal{C}|$ different class channels, its mIoU with respect to the ground truth is represented as the following:

$$\frac{1}{|\mathcal{C}|} \sum_{c \in \mathcal{C}} \Phi(A_c), \tag{S4}$$

where $\Phi : 2^\mathcal{I} \to \mathbb{R}$ is the IoU function that calculates the per-class IoU of a segmentation map, and $A_c := \{p \,|\, p \in \mathcal{I} \text{ and the predicted label of } p \text{ is } c\} \in 2^\mathcal{I}$ is the segmentation map of a class $c \in \mathcal{C}$. The IoU with respect to the ground truth of a class is calculated by dividing the overlapped regions between the predicted segmentation and the ground truth, by the union of them. Therefore, given the ground truth segmentation map $A_c^{gt} := \{p \,|\, p \in \mathcal{I} \text{ and the ground truth label of } p \text{ is } c\} \in 2^\mathcal{I}$ of a class $c$, $\Phi(A_c)$ can be represented as the following:

$$\Phi(A_c) = \frac{|A_c^{gt} \cap A_c|}{|A_c^{gt} \cup A_c|}. \tag{S5}$$

### S2.2 Differences between Channel-Wise Fusion and Pixel-Wise Fusion

Channel-wise fusion ($f^{Channel}$) differs from pixel-wise fusion ($f^{Pixel}$) in that the mIoU's of the fused pseudo labels from $f^{Channel}$ are dependent on two additional factors: (1) the fusion policy $\pi$, and (2) the conflict-resolving mechanism that assigns the value of $\varepsilon$. For (1), since the fusion policy $\pi : \mathcal{C} \to \mathcal{T}$ is a mapping function that assigns each $c \in \mathcal{C}$ to a teacher model $t \in \mathcal{T}$, there may exist $|\mathcal{T}|^{|\mathcal{C}|}$ possible mappings for $\pi$. For (2), since the conflict-resolving mechanism assigns a class label $\varepsilon \in \mathcal{C}_p^\pi \cup \{c_0\}$ to each of the pixels in $A_o^\pi$ given a $\pi$, there may exist $|\mathcal{C}_p^\pi \cup \{c_0\}|^{|A_o^\pi|}$ possible fusion outcomes. In order to examine how these factors can impact the mIoU's of the fused pseudo labels generated by $f^{Channel}$, in the following section, we analyze the scenarios when the effectiveness of $f^{Channel}$ is maximized and when it is minimized.

### S2.3 Influences of the Conflict-Resolving Mechanism

The conflict-resolving mechanism is a method that assigns a class label for $\varepsilon \in \mathcal{C}_p^\pi \cup \{c_0\}$. Based on the definition of IoU and the formulation of $f^{Channel}$, the IoU's of the fused pseudo labels generated using $f^{Channel}$ w.r.t. $y_{tgt}$ for a given class $c \in \mathcal{C}$ and an arbitrary fusion policy $\pi$ are maximized when the following conditions are met. An illustration of these conditions is plotted in Fig. S1 (a).

- **Condition a.1:** The conflict-resolving mechanism assigns class label $c$ to the pixels under the area $A_{o_1,c}^\pi := A_{o,c}^\pi \cap A_c^{gt}$, where $A_{o,c}^\pi := A_o^\pi \cap A_c^\pi$ is the overlapped area of class $c$ and the other class(es).

15

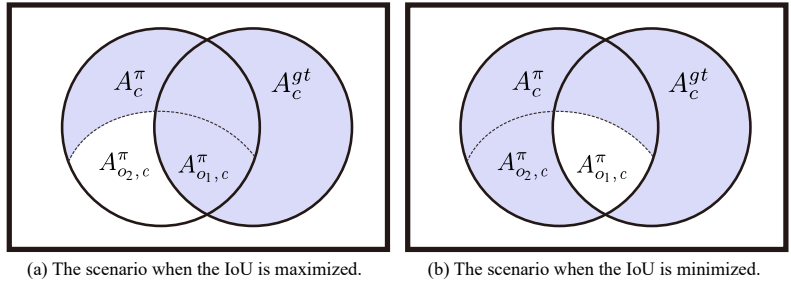

(a) The scenario when the IoU is maximized.    (b) The scenario when the IoU is minimized.

Figure S1: An illustration of the scenarios under which the IoU's of the fused pseudo label generated using $f^{Channel}$ can be maximized or minimized.

- **Condition a.2:** The conflict-resolving mechanism assigns class label $c' \in \mathcal{C}_p^\pi \cup \{c_0\}$, $c' \neq c$ to the pixels under the area $A_{o_2,c}^\pi := A_{o,c}^\pi \setminus A_c^{gt}$.

Under such conditions, the IoU w.r.t. the target domain ground truth ($y_{tgt}$) for class $c$ is given by:

$$\tilde{\Phi}^{*(c,\pi(c))} := \frac{|A_c^{gt} \cap (A_c^\pi \setminus A_{o_2,c}^\pi)|}{|A_c^{gt} \cup (A_c^\pi \setminus A_{o_2,c}^\pi)|} = \frac{|A_c^{gt} \cap A_c^\pi|}{|A_c^{gt} \cup (A_c^\pi \setminus A_{o_2,c}^\pi)|}. \tag{S6}$$

Eq. (S6) suggests that $\tilde{\Phi}^{*(c,\pi(c))} \geq \Phi^{(c,\pi(c))}, \forall c \in \mathcal{C}$, where $\Phi^{(c,\pi(c))} := \Phi(A_c^\pi) = \frac{|A_c^{gt} \cap A_c^\pi|}{|A_c^{gt} \cup A_c^\pi|}$ is the IoU w.r.t. $y_{tgt}$ for class $c$ before applying the conflict resolving mechanism.

In contrast, the IoU's of the fused pseudo labels generated using $f^{Channel}$ w.r.t. $y_{tgt}$ for a given class $c \in \mathcal{C}$ and an arbitrary fusion policy $\pi$ are minimized when the following conditions are met. An illustration of these conditions is plotted in Fig. S1 (b).

- **Condition b.1:** The conflict-resolving mechanism assigns the class label $c' \in \mathcal{C}_p^\pi \cup \{c_0\}$, $c' \neq c$ to pixels under the area $A_{o_1,c}^\pi$.

- **Condition b.2:** The conflict-resolving mechanism assigns the class label $c$ to pixels under the area $A_{o_2,c}^\pi$.

Under such conditions, the IoU w.r.t. the target domain ground truth ($y_{tgt}$) for class $c$ is given by:

$$\tilde{\Phi}'^{(c,\pi(c))} := \frac{|A_c^{gt} \cap (A_c^\pi \setminus A_{o_1,c}^\pi)|}{|A_c^{gt} \cup (A_c^\pi \setminus A_{o_1,c}^\pi)|}. \tag{S7}$$

Based on the definition in Eq. (S7), the inequality $\tilde{\Phi}'^{(c,\pi(c))} \leq \Phi^{(c,\pi(c))}, \forall c \in \mathcal{C}$ holds.

**Proposition S1.** $\forall c \in \mathcal{C}, \tilde{\Phi}^{*(c,\pi(c))} = \Phi^{(c,\pi(c))} = \tilde{\Phi}'^{(c,\pi(c))}$ if and only if $|A_o^\pi| = 0$.

*Proof.* ($\Rightarrow$) $\forall c \in \mathcal{C}, \tilde{\Phi}^{*(c,\pi(c))} = \Phi^{(c,\pi(c))} = \tilde{\Phi}'^{(c,\pi(c))}$, the following equality holds:

$$\forall c \in \mathcal{C}, \tilde{\Phi}^{*(c,\pi(c))} = \frac{|A_c^{gt} \cap A_c^\pi|}{|A_c^{gt} \cup (A_c^\pi \setminus A_{o_2,c}^\pi)|} = \frac{|A_c^{gt} \cap (A_c^\pi \setminus A_{o_1,c}^\pi)|}{|A_c^{gt} \cup (A_c^\pi \setminus A_{o_1,c}^\pi)|} = \tilde{\Phi}'^{(c,\pi(c))}$$

$$\Rightarrow \forall c \in \mathcal{C}, |A_c^{gt} \cap A_c^\pi||A_c^{gt} \cup (A_c^\pi \setminus A_{o_1,c}^\pi)| = |A_c^{gt} \cap (A_c^\pi \setminus A_{o_1,c}^\pi)||A_c^{gt} \cup (A_c^\pi \setminus A_{o_2,c}^\pi)|$$

16

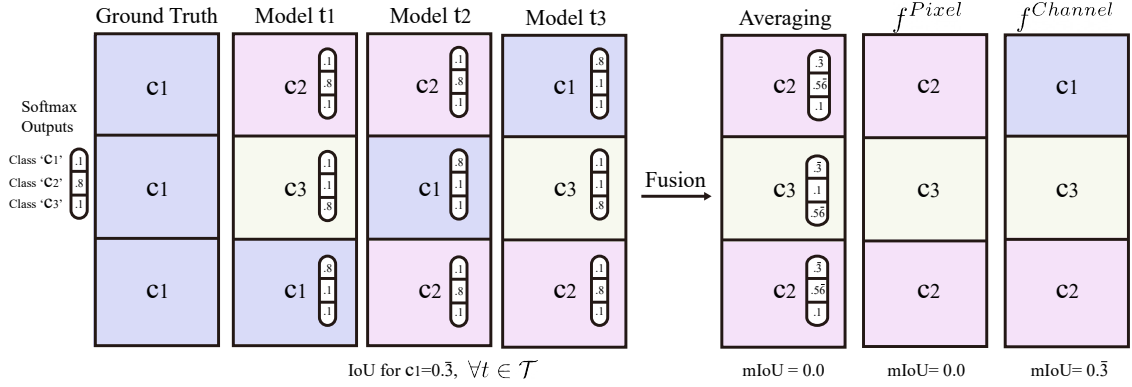

Figure S2: An illustration of the counter example described in **Proposition 2**. Each column in the figure represents a segmentation map with three pixels. The notations '$t_1$', '$t_2$', '$t_3$' represent three different teacher models in $\mathcal{T}$, and '$c_1$', '$c_2$', '$c_3$' represent the class labels in $\mathcal{C}$. The small stripes with three digits indicate the softmax outputs for those classes. In the illustrated example, the IoU's of class '$c_1$' for the models '$t_1$','$t_2$','$t_3$' are all greater than a positive constant $\alpha$ (e.g., 0.3). However, after fusion, the mIoU's of the fused results generated using averaging or $f^{Pixel}$ are equal to zero. On the contrary, the mIoU of the fused results generated by $f^{Channel}$ is greater than $\frac{n\alpha}{|\mathcal{C}|}$ (e.g., $\frac{1\times 0.3}{3}$), when a constant fusion policy $\pi(c) = t_3, \forall c \in \{c_1, c_2, c_3\}$ is adopted.

Based on the definition of $A^{\pi}_{o_1,c}$ and $A^{\pi}_{o_2,c}$, the above equation can be re-formulated as follows:

$$\forall c \in \mathcal{C}, |A^{gt}_c \cap A^{\pi}_c|(|A^{gt}_c \cup A^{\pi}_c| - |A^{\pi}_{o_1,c}|) = (|A^{gt}_c \cap A^{\pi}_c| - |A^{\pi}_{o_1,c}|)(|A^{gt}_c \cup A^{\pi}_c| - |A^{\pi}_{o_2,c}|)$$

$$\Rightarrow \forall c \in \mathcal{C}, |A^{\pi}_{o_1,c}||A^{\pi}_{o_2,c}| = |A^{gt}_c \cap A^{\pi}_c||A^{\pi}_{o_2,c}| + |A^{\pi}_{o_1,c}|(|A^{gt}_c \cup A^{\pi}_c| - |A^{gt}_c \cap A^{\pi}_c|)$$

Since $|A^{gt}_c \cap A^{\pi}_c| > |A^{\pi}_{o_1,c}|$ and $(|A^{gt}_c \cup A^{\pi}_c| - |A^{gt}_c \cap A^{\pi}_c|) > 0$, $|A^{\pi}_{o_1,c}| = |A^{\pi}_{o_2,c}| = 0, \forall c \in \mathcal{C}$. This implies $|A^{\pi}_o| = 0$, as $A^{\pi}_o = \bigcup_{c \in \mathcal{C}} A^{\pi}_{o,c} = \bigcup_{c \in \mathcal{C}} (A^{\pi}_{o_1,c} \cup A^{\pi}_{o_2,c})$.

($\Leftarrow$) If $|A^{\pi}_o| = 0$, then $\forall c \in \mathcal{C}, |A^{\pi}_{o_1,c}| = |A^{\pi}_{o_2,c}| = 0$. This implies the following:

$$\forall c \in \mathcal{C}, \tilde{\Phi}^{*(c,\pi(c))} = \frac{|A^{gt}_c \cap A^{\pi}_c|}{|A^{gt}_c \cup (A^{\pi}_c \setminus A^{\pi}_{o_2,c})|} = \frac{|A^{gt}_c \cap A^{\pi}_c|}{|A^{gt}_c \cup A^{\pi}_c|} = \frac{|A^{gt}_c \cap (A^{\pi}_c \setminus A^{\pi}_{o_1,c})|}{|A^{gt}_c \cup (A^{\pi}_c \setminus A^{\pi}_{o_1,c})|} = \tilde{\Phi}^{'(c,\pi(c))}.$$

Therefore, the equality $\tilde{\Phi}^{*(c,\pi(c))} = \Phi^{(c,\pi(c))} = \tilde{\Phi}^{'(c,\pi(c))}, \forall c \in \mathcal{C}$ holds. This also implies that, under such a condition, the IoU $\tilde{\Phi}^{(c,\pi(c))}$ of the fused results achieved by $f^{Channel}$ is solely determined by the fusion policy $\pi$. □

### S2.4 Proofs for the Propositions in the Main Manuscript

In this section, we provide proofs for the two propositions in Section 4.2.3 of the main manuscript based on the discussions in Section S2.3.

**Proposition 1.** Consider an arbitrary fusion policy $\pi$. Given a constant $\alpha \in (0,1)$ and classes $c_1, ..., c_n \in \mathcal{C}$. If $\Phi^{(c_i,t)} \geq \alpha, \forall i \in \{1,...,n\}, \forall t \in \mathcal{T}$ and $|A^{\pi}_o| = 0$, we have:

$$\text{mIoU} = \frac{1}{|\mathcal{C}|} \sum_{c \in \mathcal{C}} \tilde{\Phi}^{(c,\pi(c))} \geq \frac{n\alpha}{|\mathcal{C}|}. \tag{S8}$$

17

*Proof.* As discussed in **Proposition S1**, given an arbitrary fusion policy $\pi$, if $|A_o^\pi| = 0$, then the IoU $\tilde{\Phi}$ of the fused results achieved by $f^{Channel}$ is solely determined by $\pi$ since $\tilde{\Phi}^{*(c,\pi(c))} = \Phi^{(c,\pi(c))} = \tilde{\Phi}'^{(c,\pi(c))}$. Therefore, $\tilde{\Phi}^{(c,\pi(c))} = \Phi^{(c,\pi(c))}$ holds for all $c \in \mathcal{C}$. If $\Phi^{(c_i,t)} \geq \alpha, i \in \{1,...,n\}, \forall t \in \mathcal{T}$, the mIoU of the fused results according to Eq. (S4) and the definition of $\pi$ can be expressed as the following:

$$\frac{1}{|\mathcal{C}|}\sum_{c\in\mathcal{C}}\tilde{\Phi}^{(c,\pi(c))} = \frac{1}{|\mathcal{C}|}\sum_{c\in\mathcal{C}}\Phi^{(c,\pi(c))} \geq \frac{1}{|\mathcal{C}|}(n\alpha + \sum_{c\in\mathcal{C}\setminus\{c_1,...,c_n\}}\Phi^{(c,\pi(c))}) \geq \frac{n\alpha}{|\mathcal{C}|}. \tag{S9}$$

As a result, the mIoU achieved by $f^{Channel}$ with any $\pi$ is ensured to be greater than or equal to $\frac{n\alpha}{|\mathcal{C}|}$.

On the other hand, the mIoU's achieved by either averaging or $f^{Pixel}$ are not guaranteed to be greater than $\frac{n\alpha}{|\mathcal{C}|}$ under the same condition (i.e., $\Phi^{(c_i,t)} \geq \alpha, i \in \{1,...,n\}, \forall t \in \mathcal{T}$). As demonstrated in the counter example in Fig. S2, the IoU's of class '$c_1$' for every teacher model '$t_1$', '$t_2$', '$t_3$' are greater than a constant $\alpha \in (0,1)$. However, the mIoU's of the fused results generated by averaging and $f^{Pixel}$ are below $\frac{n\alpha}{|\mathcal{C}|}$. □

**Proposition 2.** Consider an optimal fusion policy $\pi^*(c) = \arg\max_{t\in\mathcal{T}}\{\Phi^{(c,t)}\}$. Assume $|A_o^{\pi^*}| = 0$, we have:
$$\text{mIoU} = \frac{1}{|\mathcal{C}|}\sum_{c\in\mathcal{C}}\tilde{\Phi}^{(c,\pi^*(c))} \geq \frac{1}{|\mathcal{C}|}\sum_{c\in\mathcal{C}}\Phi^{(c,t)}, \forall t \in \mathcal{T}. \tag{S10}$$

*Proof.* As discussed in **Proposition S1**, given an arbitrary fusion policy $\pi$, if $|A_o^\pi| = 0$, then the IoU $\tilde{\Phi}$ of the fused results achieved by $f^{Channel}$ is solely determined by $\pi$ since $\tilde{\Phi}^{*(c,\pi(c))} = \Phi^{(c,\pi(c))} = \tilde{\Phi}'^{(c,\pi(c))}$. Therefore, $\tilde{\Phi}^{(c,\pi(c))} = \Phi^{(c,\pi(c))}$ holds for all $c \in \mathcal{C}$. Under such a condition, the optimal IoU's for every class can be reached by following a policy $\pi^*(c) = \arg\max_{t\in\mathcal{T}}\{\Phi^{(c,t)}\}$. Such a policy is a greedy one that selects $t \in \mathcal{T}$ to maximize the target domain per-class IoU's $\Phi^{(c,t)}$ w.r.t. $y_{tgt}$ for all $c \in \mathcal{C}$. This suggests that the inequality Eq. (S10) holds for $t \in \mathcal{T}$, since:

$$\frac{1}{|\mathcal{C}|}\sum_{c\in\mathcal{C}}\tilde{\Phi}^{(c,\pi^*(c))} = \frac{1}{|\mathcal{C}|}\sum_{c\in\mathcal{C}}\Phi^{(c,\pi^*(c))} \geq \frac{1}{|\mathcal{C}|}\sum_{c\in\mathcal{C}}\Phi^{(c,t)}, \forall t \in \mathcal{T}. \tag{S11}$$

□

|  | Hyperparameter Settings |
| --- | --- |
| | CBST [3] |
| Learning Rate | $1 \times 10^{-4}$ |
| Weight Decay Factor | $5 \times 10^{-3}$ |
| Momentum | 0.9 |
| Batch Size | 2 |
| Epochs | 6 with early stopping |
| Image Crop Size | $500 \times 500$ |
| Data Augmentation | Random Multi-scale Resizing (0.7∼1.3) and Horizontal Flip |
| Class Balancing Maximum Weighting | 7 |
| | MRKLD [4] |
| Learning Rate (Phase 1) | $1 \times 10^{-3}$ |
| Learning Rate (Phase 2) | $1 \times 10^{-4}$ |
| Weight Decay Factor | $5 \times 10^{-4}$ |
| Momentum | 0.9 |
| Batch Size | 32 |
| Epochs | 6 with early stopping |
| Image Crop Size | $500 \times 500$ |
| Data Augmentation | Random Cropping, Multi-scale Resizing (0.7∼1.3) and Horizontal Flip |
| | R-MRNet [5] |
| Learning Rate | $1 \times 10^{-4}$ |
| Weight Decay Factor | $5 \times 10^{-3}$ |
| Momentum | 0.9 |
| Dropout Rate | 0.5 |
| Batch Size | 9 |
| Epochs | 35 with early stopping |
| Image Crop Size | $512 \times 256$ |
| Data Augmentation | Random Cropping, Multi-scale Resizing (0.8∼1.2) and Horizontal Flip |
| Inference Re-weighting Factor ($\alpha$) | 1 |
| Inference Re-weighting Factor ($\beta$) | 0.5 |
| | DACS [6] |
| Learning Rate | $2.5 \times 10^{-4}$ |
| Weight Decay Factor | $5 \times 10^{-4}$ |
| Momentum | 0.9 |
| Batch Size | 2 (For Both the Source and the Target Domain) |
| Epochs | 80 with early stopping |
| Image Crop Size | $512 \times 512$ |
| Data Augmentation | Random Cropping |
| | EnD [7] and EnD$^2$ [8] |
| Learning Rate | $2.5 \times 10^{-4}$ |
| Weight Decay Factor | $5 \times 10^{-3}$ |
| Momentum | 0.9 |
| Batch Size | 10 |
| Epochs | 35 with early stopping |
| Image Crop Size | Original Image Size ($1024 \times 2048$ For Cityscapes) |
| Data Augmentation | Random Horizontal Flip |
| Temperature ($T$) | 1 |
| | Ours |
| Learning Rate | $2.5 \times 10^{-4}$ |
| Weight Decay Factor | $5 \times 10^{-3}$ |
| Momentum | 0.9 |
| Batch Size | 10 |
| Epochs | 35 with early stopping |
| Image Crop Size | Original Image Size ($1024 \times 2048$ For Cityscapes) |
| Data Augmentation | Random Horizontal Flip |
| Kernel Size ($\kappa$) | 13 |

Table S2: A summary of the hyperparameters used in the proposed method and the baseline methods.

# S3 A Detailed Training Guide for Reproduction

In this section, we provide a detailed training guide for reproducing our work. In Section S3.1, we offer the pseudo code as well as the link to the source code for training the proposed framework. Then, in Section S3.2, we summarize the hyper-parameters for training the proposed framework and the baselines.

## S3.1 Pseudo Code and Source Code

The pseudo code for training the proposed framework is presented in Algorithm S1. For more details about the source codes, please refer to the GitHub repository: https://github.com/Chao-Chen-Hao/Rethinking-EnD-SegUDA.

---

**Algorithm S1** The Proposed Ensemble-Distillation Method

---

1: **Input:** Ensemble $\mathcal{T}$, and dataset $\mathcal{D}_{tgt}$
2: **Output:** Student model $m_\theta$
   // Certainty-Aware Policy Selection Strategy
3: Split $\mathcal{D}_{tgt}$ into $\mathcal{D}_{tgt}^{train}$ and $\mathcal{D}_{tgt}^{val}$
4: **for** $t \in \mathcal{T}$ **do**
5:   Initialize the weights of a student model $m_\theta$.
6:   Sample $x_{tgt}$ from $\mathcal{D}_{tgt}^{train}$, and generate the fused pseudo labels $\tilde{y}^{(p,c)}$ using $f^{Channel}$ with the constant policy $\forall c \in \mathcal{C}, \pi^{const}(c) = t$.
7:   Train $m_\theta$ with the loss in Eq. (9) in the manuscript.
8:   Evaluate the average per-class output certainty values $\rho^{(c,t)}$ of $m_\theta$ with instances in $\mathcal{D}_{tgt}^{val}$.
9: **end for**
   // Ensemble-Distillation
10: Initialize the weights of a student model $m_\theta$.
11: Sample $x_{tgt}$ from $\mathcal{D}_{tgt}$, and generate the fused pseudo labels $\tilde{y}^{(p,c)}$ using $f^{Channel}$ with $\pi$ selected based on Eq. (8) in the manuscript.
12: Train $m_\theta$ with the loss in Eq. (9) in the manuscript.

---

## S3.2 Detailed Hyper-Parameter Settings

The detailed hyperparameters for training each of the teacher models in $\mathcal{T}$, EnD [7], EnD$^2$ [8], and the proposed framework are summarized in Table S2.

# S4 Additional Experimental Results

In this section, we report the additional experimental results and provide discussions on them. We first demonstrate the performance of the proposed framework under different backbone settings in Section S4.1. Next, we showcase the reproducibility and the stability of the proposed framework in Section S4.2. Finally, we present some additional visualized results of our framework in Section S4.3.

## S4.1 A Comparison of the Backbone of the Student Model

Table S3 compares the performance of our framework using different backbone architectures in the student model. The first, second, and third columns correspond to the backbone architectures, the number of trainable parameters, and the average inference speed (denoted as *IS*), respectively. The column 'Before Distillation' denotes the mIoU of the fused pseudo labels generated by $f^{Channel}$. The column 'After Distillation' refers to the student model's performance after being trained with the fused pseudo labels. As suggested in [9],

|  | | | Before Distillation | After Distillation | | Oracle |
|---|---|---|---|---|---|---|
| Model (Backbone) | Parameters | IS | mIoU (train) | mIoU (train) | mIoU (val) | mIoU (val) |
| Deeplabv2 (ResNet-101) | 43.9 M | 33.1 ms |  | 51.76 | 52.29 | 62.54 |
| Deeplabv2 (DRN-D-54) | 35.6 M | 18.8 ms |  | 54.14 | 55.25 | 70.25 |
| Deeplabv2 (MobileNetV2) | 2.0 M | 16.5 ms | 56.31 | 48.83 | 50.98 | 60.18 |
| Deeplabv3+ (ResNet-101) | 59.3 M | 35.1 ms |  | 51.71 | 54.75 | 67.43 |
| Deeplabv3+ (DRN-D-54) | 40.7 M | 22.1 ms |  | 55.46 | 57.98 | 72.32 |
| Deeplabv3+ (MobileNetV2) | 5.8 M | 20.9 ms |  | 52.75 | 54.00 | 65.25 |

Table S3: A comparison of the performance of the proposed framework using different backbone architectures (ResNet-101, DRN-D-54, and MobileNetV2) in the student model. The numerical results are evaluated on the GTA5→Cityscapes benchmark. The inference speed is derived based on the average over 500 inferences. '*IS*' denotes the inference speed evaluated on an NVIDIA GTX TITAN V GPU. 'mIoU (train)' refers to the mIoU evaluated on the training set of Cityscapes, which includes 2975 instances. 'mIoU (val)' represents the mIoU evaluated on the validation set of Cityscapes, which includes 500 instances. The column 'Before Distillation' refers to the mIoU of the fused pseudo labels generated by $f^{Channel}$, while 'After Distillation' represents the mIoU of the student's predictions. 'Oracle' refers to the experimental setting that the student is trained directly with $y_{tgt}$ in the training set of Cityscapes and evaluated on the validation set of Cityscapes.

| GTA5 → Cityscapes | | | | | | | | | | | | | | | | | | | |
|---|---|---|---|---|---|---|---|---|---|---|---|---|---|---|---|---|---|---|---|
| Model (Backbone) | Road | SideW | Build | Wall | Fence | Pole | Light | Sign | Veg | Terrain | Sky | Person | Rider | Car | Truck | Bus | Train | Motor | Bike | mIOU |
| Deeplabv2 (ResNet-101) | 92.89 ± 0.10 | 55.61 ± 1.15 | 84.42 ± 0.10 | 41.09 ± 0.75 | 36.53 ± 0.45 | 26.16 ± 0.11 | 37.39 ± 0.15 | 46.14 ± 0.10 | 82.82 ± 0.07 | 44.68 ± 0.34 | 81.96 ± 0.30 | 56.27 ± 0.21 | 32.94 ± 0.38 | 83.27 ± 0.10 | 54.82 ± 1.25 | 46.59 ± 0.46 | 0.00 ± 0.00 | 34.27 ± 0.46 | 50.72 ± 0.54 | 52.07 ± 0.24 |
| Deeplabv3+ (MobileNetV2) | 93.32 ± 0.06 | 59.17 ± 1.05 | 86.20 ± 0.17 | 33.58 ± 1.19 | 37.85 ± 1.21 | 37.45 ± 0.32 | 43.67 ± 0.43 | 52.36 ± 0.79 | 86.34 ± 0.12 | 43.54 ± 1.11 | 86.34 ± 0.45 | 62.81 ± 0.26 | 34.53 ± 0.42 | 86.72 ± 0.86 | 46.07 ± 2.02 | 45.81 ± 1.18 | 0.00 ± 0.00 | 32.00 ± 3.60 | 53.74 ± 3.43 | 53.63 ± 0.45 |
| Deeplabv3+ (DRN-D-54) | 94.50 ± 0.22 | 61.58 ± 1.55 | 87.91 ± 0.15 | 35.87 ± 0.85 | 39.68 ± 0.89 | 40.74 ± 0.35 | 48.90 ± 0.67 | 55.13 ± 0.44 | 88.20 ± 0.05 | 48.93 ± 0.47 | 88.57 ± 0.39 | 67.06 ± 0.53 | 38.78 ± 1.12 | 89.26 ± 0.20 | 55.00 ± 2.74 | 50.48 ± 1.25 | 0.02 ± 0.06 | 40.03 ± 0.95 | 54.91 ± 1.20 | 57.13 ± 0.28 |
| SYNTHIA → Cityscapes | | | | | | | | | | | | | | | | | | | |
| Model (Backbone) | Road | SideW | Build | Wall | Fence | Pole | Light | Sign | Veg | Terrain | Sky | Person | Rider | Car | Truck | Bus | Train | Motor | Bike | mIOU |
| Deeplabv2 (ResNet-101) | 87.83 ± 0.04 | 43.42 ± 0.31 | 81.17 ± 0.11 | 18.85 ± 0.37 | 3.69 ± 0.29 | 26.07 ± 0.10 | 27.65 ± 0.86 | 34.05 ± 0.27 | 80.78 ± 0.10 | - | 82.60 ± 0.19 | 54.82 ± 0.33 | 18.78 ± 0.16 | 83.63 ± 0.16 | - | 46.09 ± 1.38 | - | 20.08 ± 0.64 | 49.05 ± 0.21 | 47.41 ± 0.15 |
| Deeplabv3+ (MobileNetV2) | 88.72 ± 0.18 | 46.91 ± 0.35 | 82.90 ± 0.16 | 18.68 ± 0.53 | 3.89 ± 0.16 | 34.4 ± 0.24 | 29.61 ± 1.18 | 36.93 ± 0.15 | 84.13 ± 0.17 | - | 88.25 ± 0.13 | 60.18 ± 0.24 | 19.35 ± 0.23 | 87.01 ± 0.24 | - | 49.01 ± 1.67 | - | 16.0 ± 2.66 | 52.30 ± 0.15 | 49.89 ± 0.26 |
| Deeplabv3+ (DRN-D-54) | 88.64 ± 0.19 | 47.04 ± 0.36 | 83.59 ± 0.08 | 19.43 ± 0.39 | 3.03 ± 0.31 | 36.11 ± 0.14 | 32.15 ± 2.57 | 37.87 ± 0.29 | 84.39 ± 0.35 | - | 87.56 ± 0.44 | 63.35 ± 0.41 | 21.12 ± 0.58 | 87.94 ± 0.20 | - | 52.58 ± 1.10 | - | 21.93 ± 1.97 | 53.76 ± 0.80 | 51.28 ± 0.13 |

Table S4: Validation of the stability and the reproducibility for the proposed framework on the GTA5 → Cityscapes and the SYNTHIA → Cityscapes benchmarks. The middle columns and the last column report the per-class IoU's and the mIoU's, respectively. Different rows correspond to different backbone configurations. Each of the numerical results are obtained from five models trained with different initial random seeds without early-stopping.

the distillation process typically requires a larger backbone to fully learn the knowledge from the teachers. However, adopting a larger backbone contradicts the core idea of ensemble-distillation, as the objective is to reduce the model size so that the computational cost at deployment time is affordable. Therefore, in our experiments, a stronger backbone 'Deeplabv3+ (DRN-D-54)' is adopted, as its number of parameters is comparable with 'Deeplabv2 (ResNet-101)' adopted by the members in $\mathcal{T}$, while performing predictions with better effectiveness. Under such a setting, it is observed that the student model is able to effectively approximate the fused pseudo labels, as mIoU's (train) only degrade slightly (0.85%) after distillation.

## S4.2 The Reproducibility and the Stability of the Proposed Framework

Table S4 demonstrates the reproducibility and the stability of the proposed ensemble-distillation framework. Each row in the table corresponds to a backbone configuration. Each of the numerical results is obtained from five models trained with different initial random seeds. From Table S4, it is observed that both the per-class IoU's and the mIoU's show only slight fluctuations in terms of their variances, indicating that the proposed method is relatively stable and thus is reproducible.

## S4.3 Visualization

Fig. S3 shows a few additional visualized results that qualitatively demonstrate the effectiveness of the proposed framework.

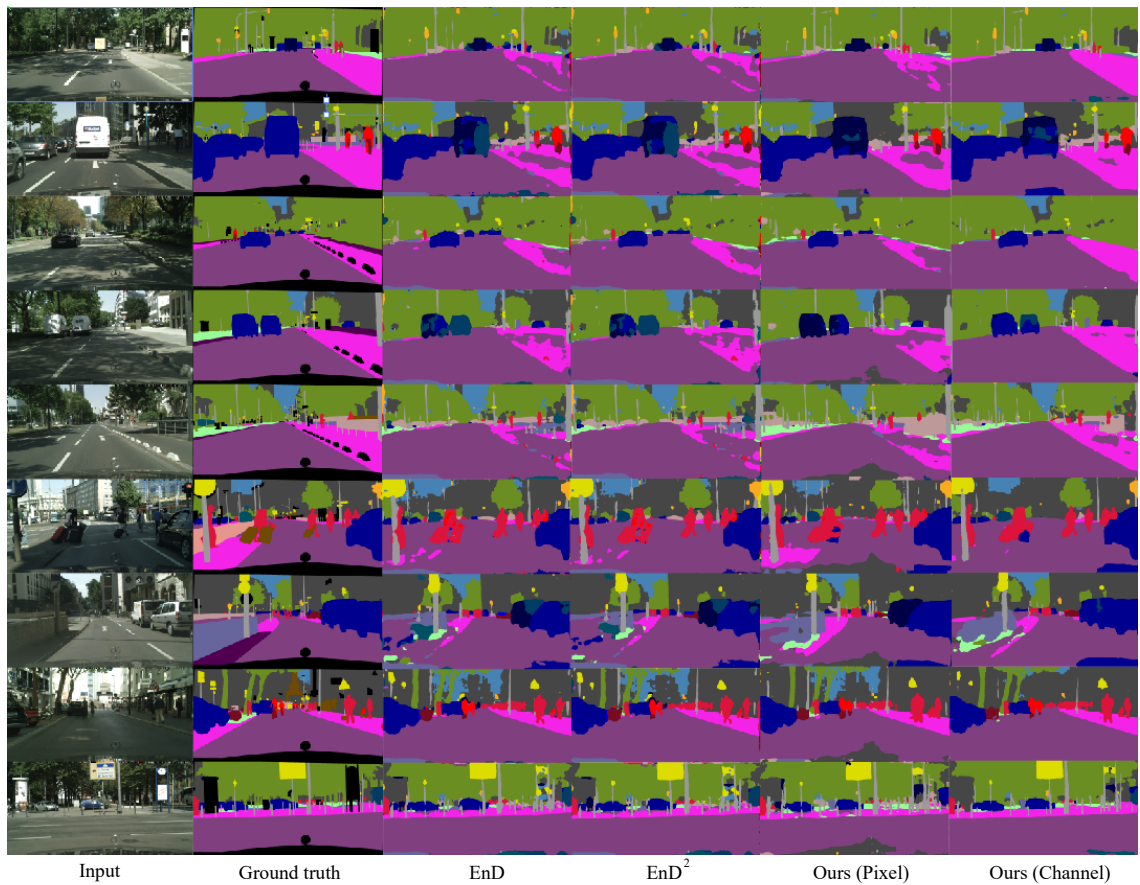

Input     Ground truth     EnD     EnD$^2$     Ours (Pixel)     Ours (Channel)

Figure S3: The visualized results evaluated on the validation set of Cityscapes. These figures are presented for qualitatively comparing the student models trained by EnD [7], EnD$^2$ [8], as well as those trained by the proposed framework with pixel-wise fusion (i.e., Ours (Pixel)) and channel-wise fusion (i.e., Ours (Channel)).



\end{document}